\documentclass[10pt,twocolumn,letterpaper]{article}

\usepackage[pagenumbers]{iccv} 

\PassOptionsToPackage{table}{xcolor}
\usepackage{xcolor}
\usepackage{multirow}
\usepackage{tabularx}
\usepackage{colortbl}
\usepackage{soul}

\definecolor{iccvblue}{rgb}{0.21,0.49,0.74}
\usepackage[pagebackref,breaklinks,colorlinks,allcolors=iccvblue]{hyperref}
\newcommand{\update}[1]{{#1}}

\def\paperID{5062} 
\def\confName{ICCV}
\def\confYear{2025}

\title{Street Gaussians without 3D Object Tracker}

\author{Ruida Zhang$^{1,2}$, Chengxi Li$^{1}$, Chenyangguang Zhang$^{1}$, Xingyu Liu$^{1}$, Haili Yuan$^{1}$, \\ Yanyan Li$^{2}$, Xiangyang Ji$^{1}$, Gim Hee Lee$^{2}$ \\
\textsuperscript{1}Tsinghua University, \textsuperscript{2}National University of Singapore
\\
{\tt\small \{zhangrd23@mails, xyji@\}.tsinghua.edu.cn, gimhee.lee@nus.edu.sg} 
}

\begin{document}
\twocolumn[{%
\renewcommand\twocolumn[1][]{#1}%
\maketitle
\vspace{-6mm}
\includegraphics[width=0.99\textwidth]{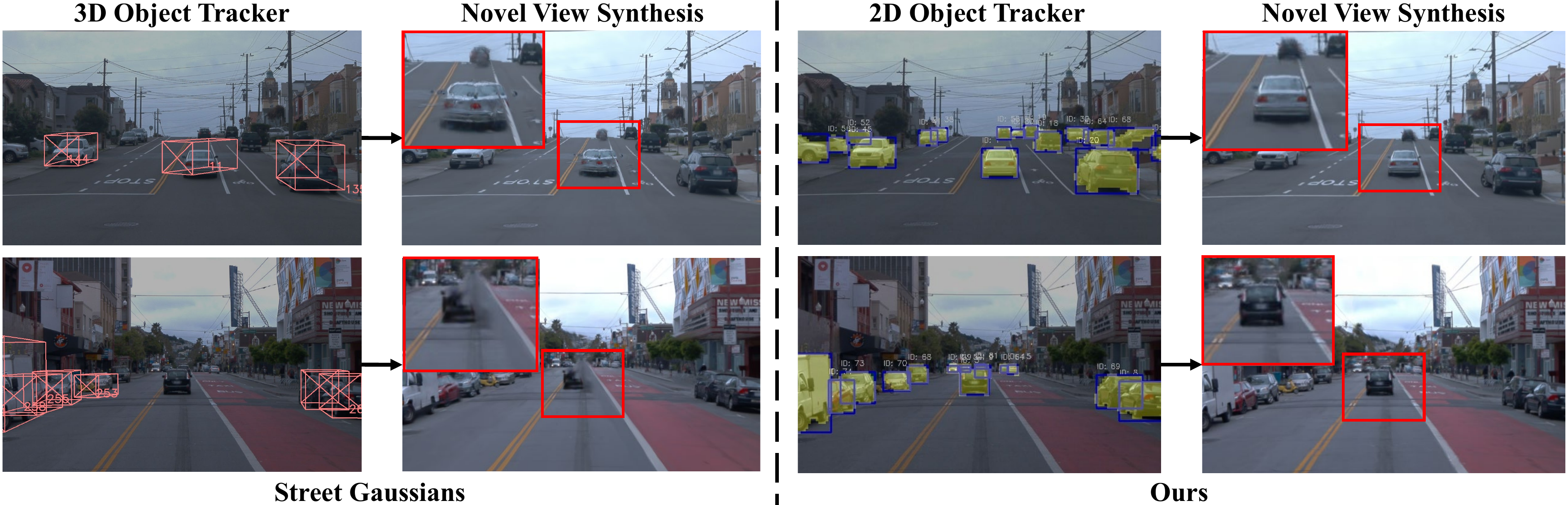}
\centering
\captionof{figure}{Comparison of 3D tracker-based Street Gaussians \cite{yan2024streetgaussian} (left) and our approach (right).
Existing methods heavily rely on object poses, but 3D trackers struggle with limited generalization \cite{soum2023mdt3d, zhang2023uni3d, eskandar2024empirical}, leading to flaws in novel view synthesis. 
In contrast, 2D foundation models show better generalization \cite{yan2023uninext,wu2024glee}. 
Our approach leverages a 2D foundation model \cite{wu2024glee} for object tracking and learns point motion within an implicit feature space to autonomously correct tracking errors, improving robustness across diverse scenes. \vspace{4mm}
}
\label{fig:teaser}
}]
\begin{abstract} 
Realistic scene reconstruction in driving scenarios poses significant challenges due to fast-moving objects. 
Most existing methods rely on labor-intensive manual labeling of object poses to reconstruct dynamic objects in canonical space and move them based on these poses during rendering. 
While some approaches attempt to use 3D object trackers to replace manual annotations, the limited generalization of 3D trackers -- caused by the scarcity of large-scale 3D datasets -- results in inferior reconstructions in real-world settings.
In contrast, 2D foundation models demonstrate strong generalization capabilities. 
To eliminate the reliance on 3D trackers and enhance robustness across diverse environments, we propose a stable object tracking module by leveraging associations from 2D deep trackers within a 3D object fusion strategy. 
We address inevitable tracking errors by further introducing a motion learning strategy in an implicit feature space that autonomously corrects trajectory errors and recovers missed detections.
\update{Experimental results on Waymo-NOTR and KITTI show that our method outperforms existing approaches.}
Our code will be released on https://lolrudy.github.io/No3DTrackSG/.

\end{abstract}    \vspace{-5mm}
\section{Introduction}
\label{sec:intro}

Modeling dynamic 3D street scenes underpins modern autonomous driving by enabling realistic, controllable simulations for tasks such as perception \cite{liu2023bevfusion, chen2023voxelnext, yin2021centerpoint, li2023pillarnext, zhou2024hugs}, prediction \cite{prediction1, prediction2, prediction3}, and motion planning \cite{plan1, plan2, plan3}. 
With the rise of end-to-end autonomous systems that require real-time sensor feedback \cite{end-to-end1, end-tu-end2, end-to-end3}, high-quality scene reconstructions have become essential for closed-loop evaluations \cite{open-loop-eval1,open-loop-eval2}, particularly to simulate critical corner cases safely and cost-effectively. 

Despite extensive efforts in achieving photo-realistic reconstruction of small-scale scenes, the unique challenges posed by the large-scale and highly dynamic nature of driving scenarios complicate effective 3D scene modeling. 
To address these challenges, most existing methods \cite{khan2024autosplat,wu2023mars,zhou2024drivinggaussian, ost2021nsg, fischer2024dynamic} rely on ground truth vehicle poses to differentiate between static background and moving vehicles. 
Typically, vehicles are reconstructed in canonical space and subsequently positioned based on known poses during rendering.
However, collecting ground-truth poses is labor-intensive, limiting the {applicability of these methods to scenes beyond the existing datasets.} 

{To eliminate 
reliance {of the systems} on ground truth object poses}, Street Gaussians \cite{yan2024streetgaussian} instead uses poses generated by 3D object trackers~\cite{wu2021castrack, wu2022casa}. 
However, by defining vehicle motion solely through these tracking poses, rendering quality becomes highly dependent on pose accuracy.
While Street Gaussians optimizes object poses during training, it struggles with detection failures and large pose errors, as shown in Fig. \ref{fig:teaser}.
Unfortunately, 3D trackers often struggle to generalize effectively across different scenarios, primarily due to the scarcity of open-source large-scale 3D datasets \cite{zhang2022det_generaliz, train_in_Germany}. 
Collecting ground-truth poses for 3D tracking is time-consuming and costly, resulting in relatively few open-source datasets for autonomous driving. 
For example, even widely-used datasets such as Waymo \cite{sun2020waymo}, Nuscenes \cite{caesar2020nuscenes}, KITTI \cite{KITTI_dataset, liao2022kitti}, Pandaset \cite{xiao2021pandaset} collectively contain fewer than 4,000 annotated scenes, limiting the diversity and scale needed for robust 3D tracking.
In contrast, 2D data is far easier to collect and annotate, leading to an abundance of 2D datasets.
The BDD100K dataset \cite{yu2018bdd100k} alone includes annotations for 100,000 driving scenes, and additional large-scale 2D perception datasets, such as COCO \cite{lin2014coco}, LVIS \cite{gupta2019lvis}, and OpenImages \cite{kuznet2020openimages}, bring the total number of annotated scenes into the millions.
This wealth of 2D data has enabled 2D foundation models to achieve strong generalization across a variety of tasks, including embedding extraction \cite{oquab2023dinov2, radford2021clip}, detection \cite{owl-vit, detr}, segmentation \cite{kirillov2023segmentanything, cheng2021mask2former}, and tracking \cite{wu2024glee}. 
{
{Nonetheless,} how to use these accurate and robust 2D trackers to {effectively} model dynamic objects {in 3D} 
{is still an} open challenge for street scene reconstruction.}

In this paper, a novel architecture is designed to achieve high-fidelity novel view synthesis performance in street scenarios for autonomous driving applications. 
Unlike previous approaches 
such as Driving Gaussian~\cite{zhou2024drivinggaussian} and Street Gaussian~\cite{yan2024streetgaussian} 
that rely on ground truth vehicle poses or poses predicted by 3D trackers, the first contribution of this work is 
a stable object tracking module.
This module leverages associations from 2D deep trackers within a 3D object fusion framework to enhance robustness and accuracy.
Specifically, we integrate 2D tracking outputs with LiDAR data to trace 
{the trajectory of each vehicle} in 3D.
We then incrementally reconstruct 
{the} point cloud {of each vehicle} {frame-by-frame} in {the} canonical space, 
estimating its pose by aligning successive frames with this canonical model.
%
This approach eliminates the reliance on 3D trackers and enhances robustness across diverse environments.

Although 2D tracking models demonstrate stronger generalization ability, tracking errors 
{still can exist}, especially under {adverse} conditions such as severe occlusions or distant objects. 
Since moving the object points solely based on the tracked object pose directly exposes any tracking errors, we aim to go beyond a straightforward reliance on tracked trajectories.
Instead, we propose to \textit{learn} point motion from the predicted trajectory, equipping the model to autonomously identify and correct tracking inaccuracies, recover missed detections, and infer motion in new time steps.
An implicit representation is essential for this purpose:
1) It enables the model to refine trajectories without being constrained by bounding box tracks, facilitating smoother and more continuous motion that can adaptively respond to changes. 
2) An implicit feature space offers versatility to move each point of an object in a different way, recognizing that vehicles 
{are not} strictly rigid objects — for example, doors can open or close.
This approach makes it possible to capture subtle, dynamic changes within objects, ultimately producing more accurate reconstructions and enhancing the robustness of novel view synthesis in challenging scenarios.

\update{
{To this end}, we leverage HexPlane representation \cite{cao2023hexplane} following 4DGS \cite{4dgs}. 
HexPlane stores motion-related features by decomposing the 4D spatial-temporal space into six 2D learnable feature planes. 
A decoder then utilizes these features to predict deformation offsets, dynamically adjusting 3D Gaussians over time.
However, 4DGS relies solely on image reconstruction loss for supervision, which is insufficient for handling rapid object motion.
When a Gaussian's initial position is far from its current location, its projection on the image may fall outside the object's actual area, preventing gradient propagation and hindering optimization.
This issue is evident in S3Gaussian \cite{s3gaussian}, which struggles with moving cars due to the lack of explicit motion supervision. 
To address this, we introduce a training strategy that supervises learned point motion using the predicted trajectory, providing explicit 3D supervision to guide HexPlane in capturing motion dynamics more accurately.
}

{Our main} contributions are summarized as follows:
\begin{itemize}
    \item We eliminate the need of 3D tracker for street scene reconstruction by introducing a robust object tracking module which leverages 2D foundation model, achieving superior generalization ability across diverse scenarios.
    \item We introduce a motion learning framework to learn from the predicted trajectory in an implicit feature space, enabling it to automatically  correct pose errors and infer motion for novel time steps.
    \item \update{We outperform existing methods on Waymo-NOTR and KITTI datasets without relying on ground truth annotations.}
\end{itemize}

\section{Related Works}
\label{sec:rw}

\subsection{3D Gaussian Splatting for Dynamic Scene}
3D Gaussian Splatting (3DGS) \cite{3dgs,li2024geogaussian} has advanced scene reconstruction by enabling high-quality, real-time rendering with 3D Gaussians and efficient splat-based rasterization, reducing computation and parameters compared to NeRF-based methods \cite{mildenhall2021nerf, barron2021mipnerf, barron2022mipnerf360, barron2023zipnerf, muller2022instant_neural} and other representations \cite{collet2015high, guo2015robust, su2020robustfusion, guo2019relightables, hu2022hvtr, li2017robust, zhang2024kpred}.
While originally designed for static scenes, 3DGS has been adapted for dynamic scenes \cite{luiten2023dynamic3dgs, yang2023realtime4d, yang2024deformable3dgs, 4dgs, stearns2024marble, liang2023gaufre,liu2024gfreedet}.
Dynamic3DGS \cite{luiten2023dynamic3dgs} directly stores information for each 3D Gaussian at every timestamp, and Yang \etal \cite{yang2023realtime4d} approximate the spatiotemporal 4D volume by optimizing 4D Gaussian primitives.
Deformable-3DGS \cite{yang2024deformable3dgs} and GauFRe \cite{liang2023gaufre} employ deformation fields to model motions, while 4DGS \cite{4dgs} introduces HexPlane representation \cite{cao2023hexplane} to store spatial-temporal features efficiently. 
Gaussian Marbles \cite{stearns2024marble} leverages isotropic ``marbles'' and a divide-and-conquer trajectory learning algorithm. 
Despite these advances, challenges persist in high-speed driving scenarios with rapid object motion.
To tackle this challenge, we leverage a robust object tracking module to guide the motion learning process.


\subsection{Street Scene Reconstruction}
Existing autonomous driving simulation engines \cite{dosovitskiy2017carla, shah2018airsim, liu2024rasim, li2019aads} face high manual effort in creating virtual environments and a lack of realism in the generated data.
{The creation of} high-fidelity simulations from driving logs is {therefore} essential for advancing closed-loop training and testing. 
Recent works \cite{yang2023unisim, xie2023s_nerf, ost2021nsg, turki2023suds, song2022towards_effi, wu2023mars, yang2023emernerf, sun2024lidarf} have continuously made improvements to NeRF \cite{mildenhall2021nerf} to model street scenes dynamically. 
Despite {the} progress, NeRF-based methods remain computationally expensive and require densely overlapping views.
Building on the effective 3DGS approach \cite{3dgs} for scene reconstruction, several 3DGS-based methods \cite{chen2023pvg,peng2025desire, zhou2024drivinggaussian,yan2024streetgaussian} have emerged.
PVG \cite{chen2023pvg} model dynamic scenarios by using periodic vibration-based temporal dynamics. 
Driving Gaussian \cite{zhou2024drivinggaussian} introduces incremental static Gaussians and composite dynamic Gaussian graph. 
Street Gaussians  \cite{yan2024streetgaussian} equips Gaussians with semantic logits, and optimize dynamic parts using tracked poses from 3D tracker. 
AutoSplat \cite{khan2024autosplat} enforces geometric constraints in road and sky regions for multi-view consistency. 
Among these approaches, most of them requires ground truth object pose \cite{chen2023pvg,zhou2024drivinggaussian,khan2024autosplat} or the 3D tracker \cite{yan2024streetgaussian}.
However, manual annotation is laborious and the 3D tracker is lack in generalization ability, which limits their applications in diverse scenarios.
In contrast, S3Gaussian \cite{s3gaussian} models dynamics in a self-supervised manner using dynamic Gaussians from 4DGS \cite{4dgs}, {despite struggling} with dynamic object modeling due to the lack of explicit motion supervision. 
We address this challenge by introducing a robust object tracking strategy based on a 2D foundation model \cite{wu2024glee} and apply motion supervision from the predicted object trajectory.

\subsection{2D and 3D trackers}
Multiple Object Tracking (MOT) aims at locating multiple objects in each frame, and establishing correspondences between them across frames within input videos \cite{3DMOT_review22, 3DMOT_review23, MOTreview21, MOTreview2021_2, bewley2016sort}.
{The reliance of most MOT methods on object detection~\cite{3DMOT_review22, 3DMOT_review23, MOTreview21, pang2022simpletrack, ding20233dmotformer, zhang2021fairmot} makes detection accuracy crucial for MOT performance.}
%
%
Recent advances in 3D object detection are promising \cite{3Ddet_survey24, 3Ddet_survey23, yin2021centerpoint, chen2023voxelnext, liu2023bevfusion, li2023pillarnext, di2022gpv, zhang2022rbp, zhang2022ssp,liu_2022_catre,zhang2024lapose, huang2025givepose,liu_2025_unopose}, but these models often generalize poorly due to limited and biased datasets \cite{soum2023mdt3d, zhang2023uni3d, eskandar2024empirical,liu2025gdrnpp}. 
Existing open-source autonomous driving datasets lack scale and show regional or environmental biases, such as vehicle density and weather conditions \cite{sun2020waymo, caesar2020nuscenes, KITTI_dataset, mao2021oncedataset}.
The main reason is that building large-scale, well-annotated multimodal datasets is costly \cite{caesar2020nuscenes, mao2021oncedataset, sanchez2023dgin3dsegment, liao2022kitti}.
In contrast, 2D image data is easier to capture and annotate, enabling extensive dataset collections \cite{che2019d2city, yu2018bdd100k, deng2009imagenet, kuznet2020openimages}. 
Visual foundation models trained on large-scale 2D data have shown strong generalization \cite{wu2024glee, cheng2021mask2former, radford2021clip, kirillov2023segmentanything, oquab2023dinov2}. 
We thus opt to use 2D object tracker \cite{wu2024glee} to locate the dynamic objects.

\begin{figure*}[t]
\centering
\includegraphics[width=0.98\textwidth]{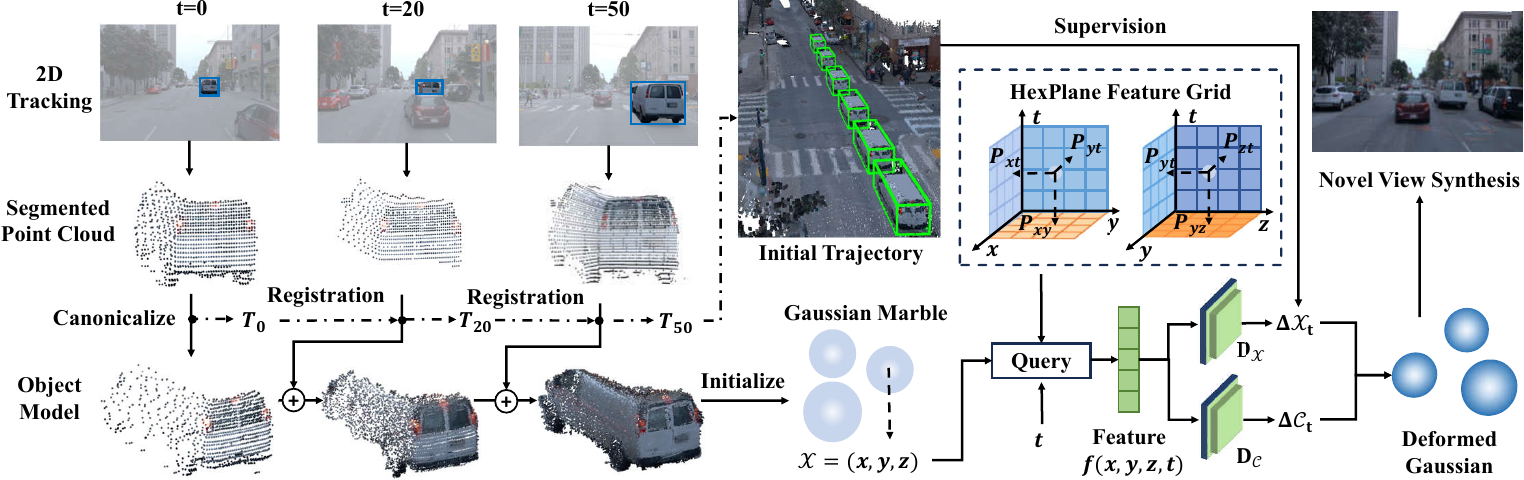}
\caption{
\textbf{{Overview of our method.}}
To overcome the limited generalization of 3D object trackers, we introduce a robust object tracking module based on a 2D object tracker \cite{wu2024glee}. 
We integrate 2D tracking with LiDAR data to segment the object's point cloud and incrementally reconstruct the object model in canonical space frame-by-frame. 
The canonical model is used to estimate object pose $\mathbf{T}_t$ and serves as the initialization for Gaussian optimization. 
We model dynamic objects with isotropic Gaussian marble \cite{stearns2024marble} to simplify optimization. 
Our approach learns point motion within the HexPlane \cite{cao2023hexplane} feature space, enabling tracking error correction and missed detection recovery. 
Decoders $D_\mathcal{X}, D_\mathcal{C}$ predict point motion $\Delta \mathcal{X}_t$ and color change $\Delta \mathcal{C}_t$ based on the HexPlane features $f(x,y,z,t)$, with the predicted trajectory providing supervision. 
Finally, novel view synthesis is performed by splatting the deformed Gaussian onto the image plane.
}
\label{fig:pipeline}
\end{figure*}

\section{{Our} Method}
\label{sec:method}

\subsection{Prerequisites: 3D Gaussian Splatting}




3D Gaussian Splatting (3DGS) \cite{3dgs} represents a 3D scene explicitly as a collection of 3D Gaussian primitives.
Each Guassian is defined by its center position $\mathcal{X} \in \mathbb{R}^3$ {and} covariance $\Sigma \in \mathbb{R}^{3 \times 3}$.
{To enforce semi-positive definiteness, the} covariance matrix is further factorized into a scaling vector $S \in \mathbb{R}^{3}$ and a rotation matrix $R \in SO(3)$ by:
\begin{equation}
\Sigma = R S S^\top R^\top
\end{equation}
Additionally, each Gaussian is described with its opacity $o \in \mathbb{R}$ and view-dependent color defined by spherical harmonic (SH) coefficient  $\mathcal{C} \in \mathbb{R}^k$, where $k$ represents numbers of SH functions. 

During rendering of novel views, differential splatting is applied to 3D Gaussians within the camera planes.
The blending of $N$ ordered points that overlap a pixel is given by the formula:
\begin{equation}
    C = \sum_{i \in N} c_i \alpha_i \prod_{j=1}^{i-1} (1 - \alpha_j), 
\end{equation}
where $\alpha_i$ and $c_i$ represents the opacity and color of the $i$th splatted Gaussian, which is computed from per-point opacity and SH coefficients (see \cite{3dgs} for details).

\subsection{Overview}
Our method takes as input multi-view images \( \mathbf{I}_t^{(j)} \) from multiple cameras {placed} around the vehicle, each indexed by time step \( t \) and camera index \( j \), along with intrinsic \( \mathbf{K}^{(j)} \) and extrinsic \( \mathbf{E}_t^{(j)} \) matrices for each view. 
Additionally, a top-mounted LiDAR provides 3D point clouds \( \mathbf{L}_t \) for each frame. 
Using this multi-sensor data, our approach reconstructs the 3D scene and synthesizes novel views from any desired camera pose and time frame, {without reliance on 3D object trackers or ground truth object trajectories}.

We present an overview of our method in Fig. \ref{fig:pipeline}.
Visual images and point clouds are obtained from the sensor setup.
For images, static and dynamic components are analyzed first.
We employ Mask2Former \cite{cheng2021mask2former} to segment the scene into static and dynamic parts.
Specifically, the dynamic part includes humans (pedestrians, cyclists, \etc) and vehicles (cars, trucks, \etc), and all the other objects are categorized as static.
We model the static part as in 3DGS \cite{3dgs} and focus on modeling dynamic objects in the following sections.
To enhance robustness, we propose an object tracking strategy by leveraging a robust 2D tracker \cite{wu2024glee} and the LiDAR point cloud, providing accurate initial point clouds and object trajectories for subsequent Gaussian optimization (
\cf Sec \ref{sec:tracking}).
We leverage an implicit representation HexPlane \cite{cao2023hexplane} following 4DGS \cite{4dgs} to learn a continuous and smooth per-point motion based on the object trajectory (
\cf Sec \ref{sec:motion}).
The optimization objective is described in Sec \ref{sec:training}.

\subsection{Tracking Cars in 3D with Robust 2D Tracker \label{sec:tracking}}

To improve the robustness of novel view synthesis for diverse scenarios, the first challenge is how to track and build vehicles in 3D based on the 2D tracking results.

We employ GLEE \cite{wu2024glee} to track all vehicles within the 2D image plane, obtaining 2D object trajectories in each camera view. 
Each trajectory includes the 2D segmentation masks of the object over a time period.
To lift 2D trajectories into 3D, we first re-project the LiDAR point clouds onto the image plane and assign points to objects based on their presence within the corresponding 2D segmentation masks {$\mathcal{M}$}.
This yields a segmented object point clouds for each camera view
{via the following function:}
\begin{equation}
       \mathcal{O}_t^{(j)} = \{\mathbf{P} | \Pi^{-1}(\mathbf{P}, \mathbf{K}^{(j)}, \mathbf{E}^{(j)}_{t}) \in \mathcal{M}, \mathbf{P}\in \mathbf{L}_t \} ,
\end{equation}
where $j$ is the camera index, $t$ is the time index, $\mathbf{P}$ is a LiDAR point and $\Pi(\cdot)$ is the re-projection function.
Given the slight misalignment between the LiDAR point cloud and RGB image, points near the edges of segmentation masks often fall outside the object boundaries.
To address this {issue}, we perform outlier removal to improve stability in subsequent reconstruction based on the point distance to the point cloud center.
We {then} associate the same objects across different camera views. Two point clouds in different viewpoints are considered 
{belonging} to the same object if they share more than 50 points in one time step.
This association process results in a set of associated partial object point cloud in different time steps. 

After obtaining the 3D partial objects in different time steps, the next step is to associate these partial point cloud into a unified and complete model. 
We initialize 
{the} 3D reconstruction {of each object} from the first frame 
which it appears.
For simplicity, we assume an object is visible between frames $0$ and $T$ and denote the object point cloud at time $t$ as $\mathcal{O}_t$.
Our goal is to obtain a temporally consistent reconstruction $\mathcal{O}$ and the object pose in each time frame.
This reconstruction is done {frame-by-frame} incrementally. 

Starting with the initial frame, we add $\mathcal{O}_0$ to $\mathcal{O}$. 
For each subsequent frame, we apply Iterative Closest Point (ICP) 
{for the alignment of} $\mathcal{O}_t$ to $\mathcal{O}$ {by} extracting the relative pose $\mathbf{T}_t$.
{The overlap between $\mathcal{O}_t$ and $\mathcal{O}$ is given by:} 
\begin{equation}
    \tau_{overlap} = \mathcal{O} \cap  \mathbf{T}_t^{-1}\mathcal{O}_t , \quad \mathbf{T}_t = \operatorname{ICP}(\mathcal{O}_t, \mathcal{O}).
\end{equation}
If ICP reveals an overlap {$\tau_{overlap}$} of more than 30\% with $\mathcal{O}$, we record the object pose $\mathbf{T}_t$ and update $\mathcal{O}$ by integrating $\mathbf{T}_t^{-1}\mathcal{O}_t$. 
If $\tau_{overlap}$ is between 10\% and 30\%, we only record the object pose $P_t$ without updating $\mathcal{O}$.
If $\tau_{overlap}$ is lower than 10\%, we assume a 2D tracking failure for that frame and discard it to filter out tracking errors.
This process is repeated for each frame using the last recorded pose as the initial guess for ICP to accelerate convergence. 
Through this approach, we reconstruct all objects and retrieve their poses in each frame, providing a robust initialization for subsequent 3D Gaussian optimization.


\subsection{Learning Point Motion \label{sec:motion}}

\renewcommand{\arraystretch}{1.25}
\begin{table*}[!ht]
    \centering
    \begin{tabularx}{\textwidth}{c |c |>{\centering\arraybackslash}X >{\centering\arraybackslash}X >{\centering\arraybackslash}X >{\centering\arraybackslash}X >{\centering\arraybackslash}X | >{\centering\arraybackslash}X >{\centering\arraybackslash}X >{\centering\arraybackslash}X >{\centering\arraybackslash}X >{\centering\arraybackslash}X}
    \hline
    \multicolumn{2}{c|}{Task} & \multicolumn{5}{c|}{Scene Reconstruction} & \multicolumn{5}{c}{Novel View Synthesis}    \\ \hline
        \small Method & \small Extra Input & \small PSNR↑ & \small SSIM↑ & \small LPIPS↓ & \small DPSNR↑ & \small DSSIM↑ & \small PSNR↑ & \small SSIM↑ & \small LPIPS↓ & \small DPSNR↑ & \small DSSIM↑ \\ \hline
        \small 3DGS \cite{3dgs} & \small N/A &25.77 & 0.833 & 0.160 & 20.26 & 0.604 & 24.46 & 0.802 & 0.170 & 18.12 & 0.521  \\ 
        \small EmerNeRF \cite{yang2023emernerf} & \small N/A & 28.16 & 0.806 & 0.228 & 24.32 & 0.682 & 25.14 & 0.747 & 0.313 & 23.49 & 0.660 \\ 
        \small S3Gaussian \cite{s3gaussian} & \small N/A & {\cellcolor{cyan!15}31.35} & {\cellcolor{cyan!15}0.911} & \cellcolor{cyan!15}0.106 & 26.02 & 0.783 & \cellcolor{cyan!15}27.44 & \cellcolor{cyan!15}0.857 & \cellcolor{cyan!15}0.137 & 22.92 & 0.680 \\ 
        \small MARS \cite{wu2023mars} & \small GT pose & 28.24 & 0.866 & 0.252 & 23.37 & 0.701 & 26.61 & 0.796 & 0.305 & 22.21 & 0.697 \\ 
        \small StreetGS \cite{yan2024streetgaussian} & \small 3D tracker & 29.17 & 0.873 & 0.138 & \cellcolor{cyan!15}27.78 & \cellcolor{cyan!15}0.818 & 26.98 & 0.838 & 0.149 & \cellcolor{cyan!15}24.62 & \cellcolor{cyan!15}0.742 \\ 
         \hline
        \small Ours & \small 2D tracker & {\cellcolor{red!15}32.56} & {\cellcolor{red!15}0.936} & {\cellcolor{red!15}0.059} & {\cellcolor{red!15}28.51} & {\cellcolor{red!15}0.868}  & {\cellcolor{red!15}28.85} & {\cellcolor{red!15}0.867} & {\cellcolor{red!15}0.088} & {\cellcolor{red!15}25.58} & {\cellcolor{red!15}0.779} 
        \\ \hline

    \end{tabularx}
    \caption{Comparison with state-of-the-art methods on Waymo-NOTR dataset. StreetGS represents Street Gaussian \cite{yan2024streetgaussian}. The \colorbox{red!15}{best} and the \colorbox{cyan!15}{second best} results are denoted by pink and blue. 
    }
    \label{tab:notr}

\end{table*}

A common approach to model object motion in prior works is directly transforming the object points using the given object pose. 
However, despite the strong generalization capabilities of 2D trackers, detection failures are inevitable in challenging scenarios, such as when objects are heavily occluded or located at great distances. 
Relying solely on object poses derived from these erroneous or missing detections can easily lead to rendering failures. 
Additionally, treating cars as rigid objects does not adequately address corner cases, such as when a car door is open or closed. 
Furthermore, these methods lack the ability to infer an object motion at arbitrary time stamps.

To address these limitations, we seek for a motion modeling approach to enhance robustness and flexibility.
Instead of explicitly using object pose as rigid transformation, we learn the per-point motion in a pre-defined feature space.
In this feature space, the object motion can be optimized through both explicit guidance and photometric loss.
Moreover, 
the object motion can be interpolated through time and space in this feature space to compensate for missing detections.
{To this end}, we leverage HexPlane representation \cite{cao2023hexplane} as in 4DGS \cite{4dgs} to efficiently capture spatial and temporal information by decomposing the 4D feature voxels into six learnable feature planes.  
This representation {satisfies} all our requirements {and is} also memory efficient.

{We use} three planes: $P_{xy}$, $P_{yz}$, and $P_{xz}$ {for the} spatial dimensions, {and another three planes} $P_{xt}$, $P_{yt}$, and $P_{zt}$ {for the} spatial-temporal features.
Additionally, the Hexplane includes multiple resolution levels 
which is formulated as:
\begin{equation}
\begin{split}
\{ P_{ij}^\rho \in \mathbb{R}^{d \times \rho{r_i} \times \rho{r_j}} | (i, j) \in \mathcal{P}, \, \rho \in \{1, 2\} \},
\end{split}
\end{equation}
where $\mathcal{P}=\{(x, y), (x, z), (y, z), (x, t), (y, t), (z, t)\}$, $d$ is the feature dimension, 
$\rho$ denotes the upsampling scale, and  $r$ is the base resolution.

Given the center position of a Gaussian $(x,y,z)$ and the time step $t$,  features in all six planes are queried and combined via a small MLP $\phi_d$ as follows:
\begin{equation}
    f(x, y, z, t) = \phi_d ( \bigcup_{\rho} \prod_{(i,j) \in \mathcal{P}} \pi ( P_{ij}^{\rho}, \psi_{ij}^{\rho} (x, y, z, t) ) ),
\end{equation}
where 
$\psi_{ij}^{\rho} (x, y, z, t)$
  projects the 4D coordinate onto the respective plane, and 
$\pi$ performs bilinear interpolation on the voxel features at each point. 
The features of all planes are combined using the Hadamard product.

Finally, two MLP decoders $D_\mathcal{X}, D_{c}$ are utilized to predict the point motion $\Delta \mathcal{X}_t$ and color change $\Delta \mathcal{C}_t $ as $\Delta \mathcal{X}_t = D_\mathcal{X}(f(x,y,z,t))$ and $\Delta \mathcal{C}_t  = D_\mathcal{C} (f(x,y,z,t))$, respectively.
The deformed 4D Gaussians are formulated as: $\mathcal{G}_t = \{\mathcal{X} + \Delta \mathcal{X}_t , \mathcal{C}  + \Delta \mathcal{C}_t , S, R, o \}$.

To allow the HexPlane feature to learn point motion from the predicted pose, 
we {define the} motion loss as:
\begin{equation}
    \mathcal{L}_\text{motion} = \operatorname{avg}_{\mathcal{X} \in \mathcal{O}}|\Delta \mathcal{X}_t - (\mathbf{T}_t \mathcal{X} - \mathcal{X}) |,
    \label{eq:motion}
\end{equation}
where $\mathcal{X}$ is the center position of a Gaussian in object $\mathcal{O}$ and $\operatorname{avg}$ is the average operator.
This loss encourages the predicted point motion $\Delta \mathcal{X}_t$ to align with the predicted pose.
We apply this loss only for the first 40\% iterations to provide a strong initial motion prior.
The loss is subsequently removed, allowing the network to adjust and potentially compensate for pose errors and detection failures.

{Following \cite{stearns2024marble}, we adopt isotropic Gaussian marbles for dynamic points to reduce degrees of freedom and simplify optimization.}
In this setup, 
{the} rotation {of each Gaussian} is represented by the identity matrix 
with identical scales across all three dimensions. 
The spherical harmonics coefficients are also limited to three dimensions, providing view-consistent color. 
This approach ensures that all deformations are captured purely by point motion and color changes, enforcing strong constraints and enhancing robustness in novel view synthesis.

\vspace{1mm}
\noindent \textbf{Ours \vs S3Gaussian.}
{Although} S3Gaussian \cite{s3gaussian} also uses Hexplane representation \cite{cao2023hexplane} and the deformation network from 4DGS \cite{4dgs}, 
our design serves a different purpose: S3Gaussian learns point motion without extra supervision, 
{while} we use 4DGS to refine motion and address detection failures.
{Consequently, S3Gaussian becomes susceptible to unsatisfactory results with rapid car motion due to the lack of explicit guidance. In contrast, our approach enables high-quality reconstruction of dynamic objects.}

\subsection{Optimization Objective \label{sec:training}}
Besides the motion loss introduced above, the loss function comprises five components 
{to collectively optimize} the scene representation and the point motion. 
The primary loss \( \mathcal{L}_{\text{rgb}} \) is an L1 loss measuring the {photometric} difference between rendered and ground truth images while \( \mathcal{L}_{\text{ssim}} \) evaluates their structural similarity. 
{
The L1 loss between the rendered depth map and the depth estimated from the LiDAR point cloud \( \mathcal{L}_{\text{depth}} \) is used to supervise the Gaussian positions.}
Following K-Planes \cite{kplanes_2023}, a grid-based total variation loss \( \mathcal{L}_{\text{tv}} \) is introduced.
Recognizing that the color of most dynamic points in the scene are unchanged, a L1 regularization loss $\mathcal{L}_\text{color-reg}$ 
is applied to the deformation network by minimizing $\Delta \mathcal{C}$.

The total loss function is thus defined as:
\begin{equation}
\begin{split}
    \mathcal{L} = &\lambda_{\text{rgb}} \mathcal{L}_{\text{rgb}} 
    + \lambda_{\text{depth}} \mathcal{L}_{\text{depth}} 
    + \lambda_{\text{ssim}} \mathcal{L}_{\text{ssim}} + \lambda_{\text{tv}} \mathcal{L}_{\text{tv}} \\
    &+ \lambda_{\text{color-reg}} \mathcal{L}_{\text{color-reg}} + \lambda_{\text{motion}} \mathcal{L}_{\text{motion}} ,
\end{split}
\end{equation}
where the weights are assigned as follows: \(\lambda_{\text{rgb}} = 1.0\), \(\lambda_{\text{depth}} = 1.0\), \(\lambda_{\text{ssim}} = 0.1\), \(\lambda_{\text{tv}} = 0.1\), \(\lambda_{\text{color-reg}} = 0.01\) and \(\lambda_{\text{motion}} = 1\).

The optimization process is divided into two stages. 
In the first stage, we render using only static Gaussians and supervise only the static regions of the images over 20,000 iterations. 
This phase focuses on reconstructing the static background and stabilizes the training process. 
In the second stage, we train all Gaussians on the whole images for 50,000 iterations.

\section{Experiments}
\label{sec:exp}

\begin{figure*}[t]
\centering
\includegraphics[width=0.95\textwidth]{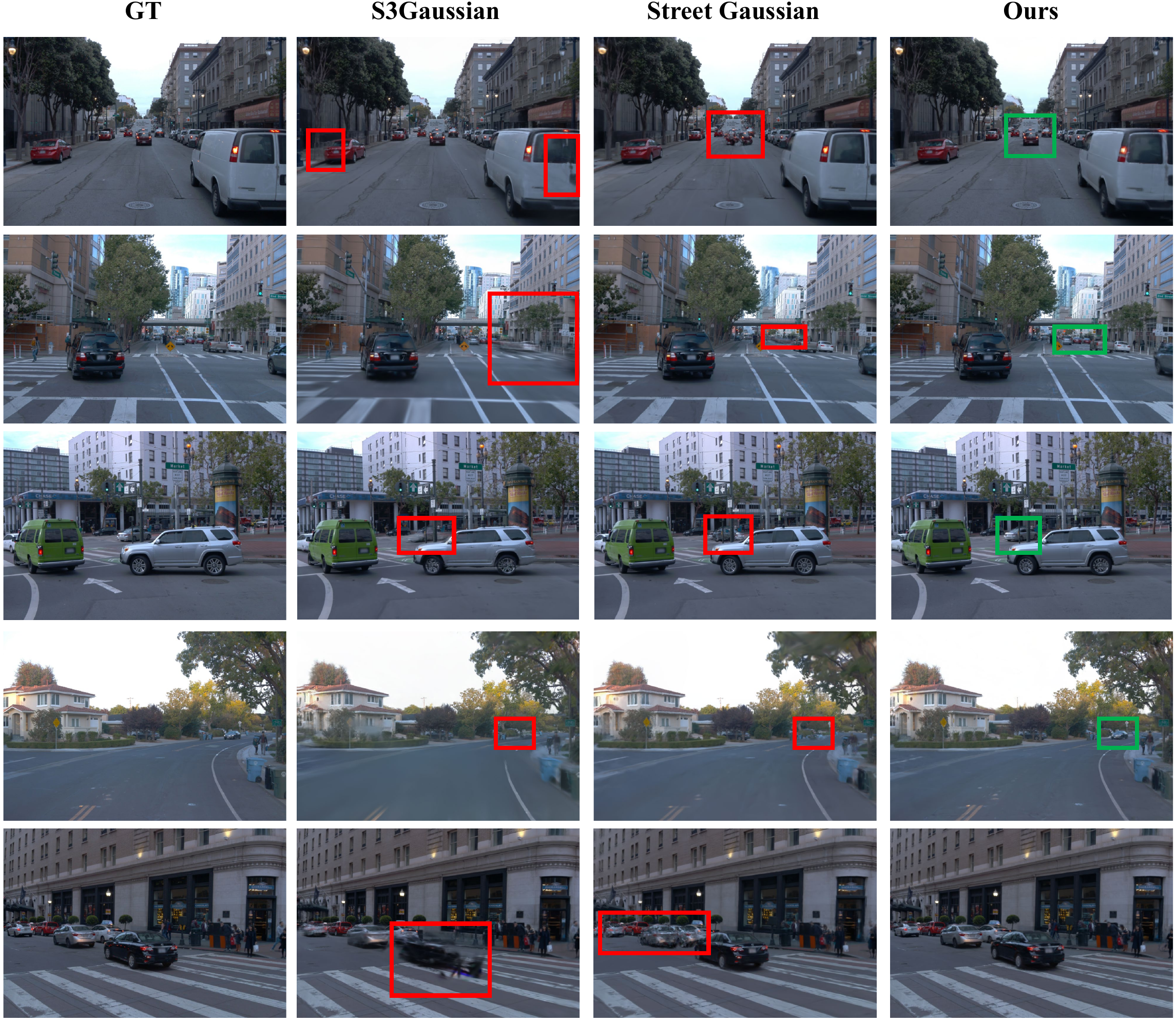}
\caption{
Qualitative comparison of novel view synthesis on NOTR dataset.
Best viewed with zoom.
}
\label{fig:vis}
\end{figure*}

\subsection{Experimental Setup}

\textbf{Datasets.}
We leverage Waymo-NOTR dataset \cite{yang2023emernerf, sun2020waymo} and KITTI benchmark \cite{KITTI_dataset, geiger2013kitti}.
The \textbf{N}eRF \textbf{O}n-\textbf{T}he-\textbf{R}oad (NOTR) dataset, introduced by EmerNeRF \cite{yang2023emernerf}, is a subset of the Waymo Open dataset \cite{sun2020waymo} and includes diverse challenging driving scenarios, such as high-speed, exposure mismatch, and various weather conditions. 
We use the dynamic32 subset, consisting of 32 dynamic scenes, for evaluating dynamic reconstruction.
Following EmerNeRF's setup \cite{yang2023emernerf}, we use three frontal camera images resized to $960 \times 640$.
\update{
For scene reconstruction, all image frames are used for training and evaluation; for novel view synthesis, every 10th time step is excluded for evaluation \cite{yang2023emernerf}.}
\update{
For the KITTI dataset, we follow 
{the} setup {of MARS} \cite{wu2023mars} 
{to use} 75\% or 50\% of the images for training
with every 4th or every 2nd frame held out for testing, respectively.
}

\vspace{1mm}
\noindent \textbf{Implementation Details.}
We use the Adam optimizer \cite{kingma2014adam} with the same learning rate schedule as in 3DGS \cite{3dgs}.
For long-sequence scene reconstruction, we follow S3Gaussian \cite{s3gaussian} and segment the scene into multiple clips.
The multi-resolution HexPlane encoder has a base resolution of 64, upsampled by factors of 2 and 4 as in \cite{4dgs}, and other hyperparameters match those in 3DGS \cite{3dgs}. 
As LiDAR points lack data for the sky region, we add a plane of points above the maximum height {of the scene} to represent it following \cite{khan2024autosplat}.
All experiments are run on a single NVIDIA RTX 3090 GPU, with training taking about 2 hours per video clip and inference speed at 100 FPS at $960 \times 640$ resolution.
  
\vspace{1mm}
\noindent \textbf{Baseline Methods.}
We compare our approach against state-of-the-art methods, including NeRF-based methods (MARS \cite{wu2023mars}, NSG \cite{ost2021nsg}, EmerNeRF \cite{yang2023emernerf}) and Gaussian-based methods (3DGS \cite{3dgs}, StreetGaussian \cite{yan2024streetgaussian}, S3Gaussian \cite{s3gaussian}). 
To ensure fair comparisons, we apply LiDAR point cloud initialization and add depth regularization to 3DGS.
For the Waymo-NOTR dataset, we borrow results of MARS \cite{wu2023mars} and EmerNeRF \cite{yang2023emernerf} from S3Gaussian \cite{s3gaussian}.\update{
We evaluate the 3D-tracker-based method, Street Gaussians \cite{yan2024streetgaussian}, using detection results from VoxelNext \cite{chen2023voxelnext} and associate the detections with SimpleTrack \cite{pang2022simpletrack}.
Our goal is to assess the {generalization ability} of 3D-tracker-based methods in scenarios without ground truth tracking labels. 
To {emulate} a real-world scenario where a 3D tracker trained on a public benchmark dataset is applied to a novel environment, we use pretrained weights from nuScenes \cite{caesar2020nuscenes}—one of the largest and most diverse 3D datasets—and evaluate the model on the Waymo-NOTR dataset.
Since NOTR is part of the training set for the Waymo perception task, using Waymo pretrained weights would introduce domain overlap and bias the evaluation, making the nuScenes weights a fairer choice.
For a fair comparison, we use pretrained 2D tracker weights from \cite{wu2024glee} for our method, which {have} not {been} trained on the Waymo or KITTI datasets.
Please refer to the supplementary material for discussion of the choice of 3D trackers.
For the KITTI dataset, we borrow {the} results of all other methods from Street Gaussians.
Please note that Street Gaussians leverages GT tracking annotations on KITTI dataset.}

\vspace{1mm}
\noindent \textbf{Metrics.} 
We leverage peak signal-to-noise ratio (PSNR), structural similarity index (SSIM) and learned perceptual image patch similarity (LPIPS) to evaluate the rendering quality.
Additionally, following EmerNeRF \cite{yang2023emernerf}, we apply DPSNR and DSSIM metrics to dynamic objects by projecting their ground truth 3D bounding boxes onto the 2D image plane and calculating pixel loss within these regions.

\subsection{Comparisons with the State-of-the-art}

\begin{table*}[t]
    \centering
    \begin{tabular}{c|c|ccc|ccc}
    \hline
    \multicolumn{2}{c|}{} & \multicolumn{3}{c|}{KITTI-75\%} & \multicolumn{3}{c}{KITTI-50\%}  \\ \hline
    Method &  Extra Input &  PSNR↑ &  SSIM↑ &  LPIPS↓ &  PSNR↑ &  SSIM↑ &  LPIPS↓
    \\ \hline
        3DGS \cite{3dgs} & N/A & 19.19 & 0.737 & 0.172 & 19.23 & 0.739 & 0.174 \\ 
        NSG \cite{ost2021nsg} & GT pose &  21.53 & 0.673 & 0.254 & 21.26 & 0.659 & 0.266  \\ 
        MARS \cite{wu2023mars} & GT pose & 24.23 & \cellcolor{cyan!15}0.845 &  0.160 & 24.00 & 0.801 &  0.164\\ 
        Street Guassians \cite{yan2024streetgaussian} & GT pose & \cellcolor{red!15}25.79 & 0.844 & \cellcolor{cyan!15}0.081 & \cellcolor{red!15}25.52 & \cellcolor{cyan!15}0.841 & \cellcolor{cyan!15}0.084 \\ \hline
        Ours & 2D tracker \cite{wu2024glee} & \cellcolor{cyan!15}25.49 & \cellcolor{red!15}0.889 & \cellcolor{red!15}0.063 & \cellcolor{cyan!15}25.11 & \cellcolor{red!15}0.877 & \cellcolor{red!15}0.067\\ \hline
    \end{tabular}
    \caption{\update{Comparison with state-of-the-art methods on KITTI dataset. The \colorbox{red!15}{best} and the \colorbox{cyan!15}{second best} results are denoted by pink and blue. }
    }
    \label{tab:kitti}
\end{table*}

On the NOTR dataset, our method outperforms all competitors across every metric, as shown in Table \ref{tab:notr}. 
Our approach sets a new state-of-the-art in both the scene reconstruction and novel view synthesis tasks. 
Specifically, our method outperforms S3Gaussian \cite{s3gaussian} by 2.66 dB in DPSNR and 0.099 in DSSIM for novel view synthesis. 
It also surpasses Street Gaussians \cite{yan2024streetgaussian} by 1.87 dB in PSNR, mainly due to the limited generalization of 3D trackers used in Street Gaussians. 
Despite optimizing object poses, Street Gaussians struggles with large pose errors and detection failures, leading to inferior performance. 
Additionally, we outperform NeRF-based methods like EmerNeRF and MARS.

We present qualitative comparison with S3Gaussian \cite{s3gaussian} and Street Gaussians \cite{yan2024streetgaussian} in Fig.~\ref{fig:vis}. Results from S3Gaussian show that using 4DGS without explicit motion guidance results in weaker performance when handling moving vehicles (\eg row 2, 3, 5). 
Street Gaussians suffers from tracking errors of 3D trackers, leading to erroneous reconstructions (\eg row 1, 3) or entirely missed objects (\eg rows 2, 4, and 5). 
In contrast, our approach performs robustly across diverse scenarios owing to our tracking strategy leveraging 2D foundation model and robust motion learning module.
\update{
Please refer to the supplementary material for \textbf{additional video results} and \textbf{editing examples}.}

\update{
The results on the KITTI dataset are presented in Table \ref{tab:kitti}. 
Our method outperforms the NeRF-based NSG and MARS across all metrics. 
In the KITTI-75\% setting, our approach achieves a 0.045 higher SSIM and a 0.018 lower LPIPS compared to Street Gaussians, although its PSNR is 0.3 dB lower. 
A similar trend is observed in the KITTI-50\% setting. 
This lower PSNR is primarily because Street Gaussians leverages labor-intensive ground truth tracking annotations, whereas our method uses a generalized 2D tracker—yet still attains comparable performance. 
}

\subsection{Ablation Studies}
\begin{table}[t]
    \centering
    \scalebox{1}{
    \begin{tabular}{c|c|cccc}
    \hline
       \small   & \small Method & \small PSNR & \small SSIM & \small DPSNR & \small DSSIM \\ \hline
        A & \small w/o track & 25.25 & 0.761 & 21.71 & 0.662 \\ 
        B & \small w/o $\mathcal{L}_{\text{motion}}$ & 28.39	 & 0.843 & 24.54 & 0.719  \\ 
        C & \small w/o iso. &  28.58 & 0.847 & 25.35 & 0.765  \\ 
        \hline
        \update{D} & \small 3D tracker & 28.02 & 0.835	& 24.61 &	0.746 \\
        \update{E} &  {D} \small w/o $\mathcal{L}_{\text{motion}}$ & 27.54 & 0.821	& 23.53 &	0.703 \\
        F & \small GT pose & 28.98 & 0.873 & 25.71 & 0.787 \\ \hline
        G & \small UNINEXT  & 28.76 & 0.864 & 25.43 & 0.772\\ \hline
        H & \small Ours   & {28.85} & {0.867}  & {25.58} & {0.779}     \\ \hline
        \end{tabular}
        }
    \caption{Ablation study of Novel View Synthesis on NOTR dataset.}
    \label{tab:ablation}
\end{table}

We conduct an ablation study on the NOTR dataset 
{and the results are} presented in Tab. \ref{tab:ablation}.

\vspace{1mm}
\noindent \textbf{Effect of object tracking module.}
In Tab.~\ref{tab:ablation} (A), we model dynamic points using 4DGS \cite{4dgs} without performing any form of tracking, and observe significant drops across all metrics.
This experiment highlight that the sole use of 4DGS does not provide sufficient accuracy for effective motion modeling in autonomous driving scenarios.

\vspace{1mm}
\noindent \textbf{Effect of motion learning strategy.}
In Tab.~\ref{tab:ablation} (B), we use the reconstruction of the tracking module for initialization but remove the motion loss introduced in Eq. \ref{eq:motion}, resulting in a 1.04 dB drop in DPSNR.
This result highlights the importance of our motion learning strategy.
A comparison with rigid-transformation-based motion modeling \cite{yan2024streetgaussian} is provided in the supplementary material.

\vspace{1mm}
\noindent \textbf{Effect of isotropic Gaussian marbles.}
In Tab.~\ref{tab:ablation} (C), we substitute isotropic Gaussian marbles with the original anisotropic Gaussian ellipsoids for dynamic points, which leads to a decrease in DPSNR and DSSIM.

\vspace{1mm}
\noindent \textbf{3D tracker / GT pose \vs Our object tracking module.}
In Tab.~\ref{tab:ablation} (D), we use the 3D tracker \cite{chen2023voxelnext, pang2022simpletrack} to reconstruct dynamic objects and supervise motion learning, resulting in a \update{0.97 dB} DPSNR decrease.
\update{Removing motion loss leads to a further 1.08 dB drop (Tab. \ref{tab:ablation} (E)), showing that our motion learning strategy compensates for tracking errors, though these errors still impact novel view synthesis, emphasizing the need for a more generalizable 2D tracker.}
Using ground-truth trajectories (Tab.~\ref{tab:ablation} (F)) shows only a small gap, confirming the effectiveness of our 2D tracker in motion learning.
See supplementary for tracking error analysis.

\vspace{1mm}
\update{
\noindent \textbf{Choice of 2D tracker.} 
To demonstrate the robustness of our method with respect to the choice of 2D tracker, we employ UNINEXT \cite{yan2023uninext} for 2D tracking. As shown in Table \ref{tab:ablation} (G), our method maintains stable performance.}

\section{Conclusion}

In this paper, we introduce a novel framework for robust dynamic 3D street scene reconstruction that eliminates the reliance on 3D object trackers.
Addressing the generalization limitations of 3D trackers, we propose a robust object tracking strategy based on a 2D foundation model.
Our framework also features a motion learning module within an implicit feature space to handle inevitable tracking errors by autonomously refining pose inaccuracies and recovering missed detections.
Experiments on the Waymo-NOTR and KITTI datasets demonstrate its adaptability and superior performance.
\update{\textbf{Limitations} and \textbf{failure cases} are discussed in the supplementary material.}

\section*{Acknowledgment}
This work was supported by the National Key R\&D Program of China under Grant 2018AAA0102801, PCL-CMCC Foundation for Science and Innovation Grant No. 2024ZY1C0040, and the National Research Foundation, Singapore, under its NRF-Investigatorship Programme (Award ID. NRF-NRFI09-0008).

{
    \small
    \bibliographystyle{ieeenat_fullname}
    \bibliography{main}

\begin{thebibliography}{108}
\providecommand{\natexlab}[1]{#1}
\providecommand{\url}[1]{\texttt{#1}}
\expandafter\ifx\csname urlstyle\endcsname\relax
  \providecommand{\doi}[1]{doi: #1}\else
  \providecommand{\doi}{doi: \begingroup \urlstyle{rm}\Url}\fi

\bibitem[Barron et~al.(2021)Barron, Mildenhall, Tancik, Hedman, Martin-Brualla, and Srinivasan]{barron2021mipnerf}
Jonathan~T Barron, Ben Mildenhall, Matthew Tancik, Peter Hedman, Ricardo Martin-Brualla, and Pratul~P Srinivasan.
\newblock Mip-nerf: A multiscale representation for anti-aliasing neural radiance fields.
\newblock In \emph{Proceedings of the IEEE/CVF international conference on computer vision}, pages 5855--5864, 2021.

\bibitem[Barron et~al.(2022)Barron, Mildenhall, Verbin, Srinivasan, and Hedman]{barron2022mipnerf360}
Jonathan~T Barron, Ben Mildenhall, Dor Verbin, Pratul~P Srinivasan, and Peter Hedman.
\newblock Mip-nerf 360: Unbounded anti-aliased neural radiance fields.
\newblock In \emph{Proceedings of the IEEE/CVF conference on computer vision and pattern recognition}, pages 5470--5479, 2022.

\bibitem[Barron et~al.(2023)Barron, Mildenhall, Verbin, Srinivasan, and Hedman]{barron2023zipnerf}
Jonathan~T Barron, Ben Mildenhall, Dor Verbin, Pratul~P Srinivasan, and Peter Hedman.
\newblock Zip-nerf: Anti-aliased grid-based neural radiance fields.
\newblock In \emph{Proceedings of the IEEE/CVF International Conference on Computer Vision}, pages 19697--19705, 2023.

\bibitem[Bashar et~al.(2022)Bashar, Islam, Hussain, Hasan, Rahman, and Kabir]{3DMOT_review22}
Mk Bashar, Samia Islam, Kashifa~Kawaakib Hussain, Md~Bakhtiar Hasan, ABM Rahman, and Md~Hasanul Kabir.
\newblock Multiple object tracking in recent times: A literature review.
\newblock \emph{arXiv preprint arXiv:2209.04796}, 2022.

\bibitem[Bewley et~al.(2016)Bewley, Ge, Ott, Ramos, and Upcroft]{bewley2016sort}
Alex Bewley, Zongyuan Ge, Lionel Ott, Fabio Ramos, and Ben Upcroft.
\newblock Simple online and realtime tracking.
\newblock In \emph{2016 IEEE international conference on image processing (ICIP)}, pages 3464--3468. IEEE, 2016.

\bibitem[Caesar et~al.(2020)Caesar, Bankiti, Lang, Vora, Liong, Xu, Krishnan, Pan, Baldan, and Beijbom]{caesar2020nuscenes}
Holger Caesar, Varun Bankiti, Alex~H Lang, Sourabh Vora, Venice~Erin Liong, Qiang Xu, Anush Krishnan, Yu Pan, Giancarlo Baldan, and Oscar Beijbom.
\newblock nuscenes: A multimodal dataset for autonomous driving.
\newblock In \emph{Proceedings of the IEEE/CVF conference on computer vision and pattern recognition}, pages 11621--11631, 2020.

\bibitem[Cao and Johnson(2023)]{cao2023hexplane}
Ang Cao and Justin Johnson.
\newblock Hexplane: A fast representation for dynamic scenes.
\newblock In \emph{Proceedings of the IEEE/CVF Conference on Computer Vision and Pattern Recognition}, pages 130--141, 2023.

\bibitem[Carion et~al.(2020)Carion, Massa, Synnaeve, Usunier, Kirillov, and Zagoruyko]{detr}
Nicolas Carion, Francisco Massa, Gabriel Synnaeve, Nicolas Usunier, Alexander Kirillov, and Sergey Zagoruyko.
\newblock End-to-end object detection with transformers.
\newblock In \emph{European conference on computer vision}, pages 213--229. Springer, 2020.

\bibitem[Che et~al.(2019)Che, Li, Li, Jiang, Shi, Zhang, Lu, Wu, Liu, and Ye]{che2019d2city}
Zhengping Che, Guangyu Li, Tracy Li, Bo Jiang, Xuefeng Shi, Xinsheng Zhang, Ying Lu, Guobin Wu, Yan Liu, and Jieping Ye.
\newblock D2-city: a large-scale dashcam video dataset of diverse traffic scenarios.
\newblock \emph{arXiv preprint arXiv:1904.01975}, 2019.

\bibitem[Chen et~al.(2023{\natexlab{a}})Chen, Gu, Jiang, Zhu, and Zhang]{chen2023pvg}
Yurui Chen, Chun Gu, Junzhe Jiang, Xiatian Zhu, and Li Zhang.
\newblock Periodic vibration gaussian: Dynamic urban scene reconstruction and real-time rendering.
\newblock \emph{arXiv preprint arXiv:2311.18561}, 2023{\natexlab{a}}.

\bibitem[Chen et~al.(2023{\natexlab{b}})Chen, Liu, Zhang, Qi, and Jia]{chen2023voxelnext}
Yukang Chen, Jianhui Liu, Xiangyu Zhang, Xiaojuan Qi, and Jiaya Jia.
\newblock Voxelnext: Fully sparse voxelnet for 3d object detection and tracking.
\newblock In \emph{Proceedings of the IEEE/CVF Conference on Computer Vision and Pattern Recognition}, pages 21674--21683, 2023{\natexlab{b}}.

\bibitem[Cheng et~al.(2022{\natexlab{a}})Cheng, Misra, Schwing, Kirillov, and Girdhar]{cheng2021mask2former}
Bowen Cheng, Ishan Misra, Alexander~G. Schwing, Alexander Kirillov, and Rohit Girdhar.
\newblock Masked-attention mask transformer for universal image segmentation.
\newblock In \emph{CVPR}, 2022{\natexlab{a}}.

\bibitem[Cheng et~al.(2022{\natexlab{b}})Cheng, Chen, Zhang, Gan, Liu, and Liu]{plan1}
Jie Cheng, Yingbing Chen, Qingwen Zhang, Lu Gan, Chengju Liu, and Ming Liu.
\newblock Real-time trajectory planning for autonomous driving with gaussian process and incremental refinement.
\newblock In \emph{2022 International Conference on Robotics and Automation (ICRA)}, pages 8999--9005. IEEE, 2022{\natexlab{b}}.

\bibitem[Cheng et~al.(2023)Cheng, Mei, and Liu]{plan2}
Jie Cheng, Xiaodong Mei, and Ming Liu.
\newblock Forecast-mae: Self-supervised pre-training for motion forecasting with masked autoencoders.
\newblock In \emph{Proceedings of the IEEE/CVF International Conference on Computer Vision}, pages 8679--8689, 2023.

\bibitem[Collet et~al.(2015)Collet, Chuang, Sweeney, Gillett, Evseev, Calabrese, Hoppe, Kirk, and Sullivan]{collet2015high}
Alvaro Collet, Ming Chuang, Pat Sweeney, Don Gillett, Dennis Evseev, David Calabrese, Hugues Hoppe, Adam Kirk, and Steve Sullivan.
\newblock High-quality streamable free-viewpoint video.
\newblock \emph{ACM Transactions on Graphics (ToG)}, 34\penalty0 (4):\penalty0 1--13, 2015.

\bibitem[Dauner et~al.(2023)Dauner, Hallgarten, Geiger, and Chitta]{plan3}
Daniel Dauner, Marcel Hallgarten, Andreas Geiger, and Kashyap Chitta.
\newblock Parting with misconceptions about learning-based vehicle motion planning.
\newblock In \emph{Conference on Robot Learning}, pages 1268--1281. PMLR, 2023.

\bibitem[Deng et~al.(2009)Deng, Dong, Socher, Li, Li, and Fei-Fei]{deng2009imagenet}
Jia Deng, Wei Dong, Richard Socher, Li-Jia Li, Kai Li, and Li Fei-Fei.
\newblock Imagenet: A large-scale hierarchical image database.
\newblock In \emph{2009 IEEE conference on computer vision and pattern recognition}, pages 248--255. Ieee, 2009.

\bibitem[Di et~al.(2022)Di, Zhang, Lou, Manhardt, Ji, Navab, and Tombari]{di2022gpv}
Yan Di, Ruida Zhang, Zhiqiang Lou, Fabian Manhardt, Xiangyang Ji, Nassir Navab, and Federico Tombari.
\newblock Gpv-pose: Category-level object pose estimation via geometry-guided point-wise voting.
\newblock In \emph{Proceedings of the IEEE/CVF Conference on Computer Vision and Pattern Recognition}, pages 6781--6791, 2022.

\bibitem[Ding et~al.(2023)Ding, Rehder, Schneider, Cordts, and Gall]{ding20233dmotformer}
Shuxiao Ding, Eike Rehder, Lukas Schneider, Marius Cordts, and Juergen Gall.
\newblock 3dmotformer: Graph transformer for online 3d multi-object tracking.
\newblock In \emph{Proceedings of the IEEE/CVF International Conference on Computer Vision}, pages 9784--9794, 2023.

\bibitem[Dosovitskiy et~al.(2017)Dosovitskiy, Ros, Codevilla, Lopez, and Koltun]{dosovitskiy2017carla}
Alexey Dosovitskiy, German Ros, Felipe Codevilla, Antonio Lopez, and Vladlen Koltun.
\newblock Carla: An open urban driving simulator.
\newblock In \emph{Conference on robot learning}, pages 1--16. PMLR, 2017.

\bibitem[Eskandar(2024)]{eskandar2024empirical}
George Eskandar.
\newblock An empirical study of the generalization ability of lidar 3d object detectors to unseen domains.
\newblock In \emph{Proceedings of the IEEE/CVF Conference on Computer Vision and Pattern Recognition}, pages 23815--23825, 2024.

\bibitem[Fischer et~al.(2024)Fischer, Kulhanek, Bulo, Porzi, Pollefeys, and Kontschieder]{fischer2024dynamic}
Tobias Fischer, Jonas Kulhanek, Samuel~Rota Bulo, Lorenzo Porzi, Marc Pollefeys, and Peter Kontschieder.
\newblock Dynamic 3d gaussian fields for urban areas.
\newblock \emph{arXiv preprint arXiv:2406.03175}, 2024.

\bibitem[Geiger et~al.(2012)Geiger, Lenz, and Urtasun]{KITTI_dataset}
Andreas Geiger, Philip Lenz, and Raquel Urtasun.
\newblock Are we ready for autonomous driving? the kitti vision benchmark suite.
\newblock In \emph{2012 IEEE conference on computer vision and pattern recognition}, pages 3354--3361. IEEE, 2012.

\bibitem[Geiger et~al.(2013)Geiger, Lenz, Stiller, and Urtasun]{geiger2013kitti}
Andreas Geiger, Philip Lenz, Christoph Stiller, and Raquel Urtasun.
\newblock Vision meets robotics: The kitti dataset.
\newblock \emph{The International Journal of Robotics Research}, 32\penalty0 (11):\penalty0 1231--1237, 2013.

\bibitem[Gu et~al.(2023)Gu, Hu, Zhang, Chen, Wang, Wang, and Zhao]{prediction3}
Junru Gu, Chenxu Hu, Tianyuan Zhang, Xuanyao Chen, Yilun Wang, Yue Wang, and Hang Zhao.
\newblock Vip3d: End-to-end visual trajectory prediction via 3d agent queries.
\newblock In \emph{Proceedings of the IEEE/CVF Conference on Computer Vision and Pattern Recognition}, pages 5496--5506, 2023.

\bibitem[Guo et~al.(2015)Guo, Xu, Wang, Liu, and Dai]{guo2015robust}
Kaiwen Guo, Feng Xu, Yangang Wang, Yebin Liu, and Qionghai Dai.
\newblock Robust non-rigid motion tracking and surface reconstruction using l0 regularization.
\newblock In \emph{Proceedings of the IEEE International Conference on Computer Vision}, pages 3083--3091, 2015.

\bibitem[Guo et~al.(2019)Guo, Lincoln, Davidson, Busch, Yu, Whalen, Harvey, Orts-Escolano, Pandey, Dourgarian, et~al.]{guo2019relightables}
Kaiwen Guo, Peter Lincoln, Philip Davidson, Jay Busch, Xueming Yu, Matt Whalen, Geoff Harvey, Sergio Orts-Escolano, Rohit Pandey, Jason Dourgarian, et~al.
\newblock The relightables: Volumetric performance capture of humans with realistic relighting.
\newblock \emph{ACM Transactions on Graphics (ToG)}, 38\penalty0 (6):\penalty0 1--19, 2019.

\bibitem[Gupta et~al.(2019)Gupta, Dollar, and Girshick]{gupta2019lvis}
Agrim Gupta, Piotr Dollar, and Ross Girshick.
\newblock Lvis: A dataset for large vocabulary instance segmentation.
\newblock In \emph{Proceedings of the IEEE/CVF conference on computer vision and pattern recognition}, pages 5356--5364, 2019.

\bibitem[Hu et~al.(2022{\natexlab{a}})Hu, Chen, Wu, Li, Yan, and Tao]{end-to-end1}
Shengchao Hu, Li Chen, Penghao Wu, Hongyang Li, Junchi Yan, and Dacheng Tao.
\newblock St-p3: End-to-end vision-based autonomous driving via spatial-temporal feature learning.
\newblock In \emph{European Conference on Computer Vision}, pages 533--549. Springer, 2022{\natexlab{a}}.

\bibitem[Hu et~al.(2022{\natexlab{b}})Hu, Yu, Zheng, Zhang, Liu, and Zwicker]{hu2022hvtr}
Tao Hu, Tao Yu, Zerong Zheng, He Zhang, Yebin Liu, and Matthias Zwicker.
\newblock Hvtr: Hybrid volumetric-textural rendering for human avatars.
\newblock In \emph{2022 International Conference on 3D Vision (3DV)}, pages 197--208. IEEE, 2022{\natexlab{b}}.

\bibitem[Hu et~al.(2023)Hu, Yang, Chen, Li, Sima, Zhu, Chai, Du, Lin, Wang, et~al.]{end-tu-end2}
Yihan Hu, Jiazhi Yang, Li Chen, Keyu Li, Chonghao Sima, Xizhou Zhu, Siqi Chai, Senyao Du, Tianwei Lin, Wenhai Wang, et~al.
\newblock Planning-oriented autonomous driving.
\newblock In \emph{Proceedings of the IEEE/CVF Conference on Computer Vision and Pattern Recognition}, pages 17853--17862, 2023.

\bibitem[Huang et~al.(2024)Huang, Wei, Zheng, An, Lu, Zhan, Tomizuka, Keutzer, and Zhang]{s3gaussian}
Nan Huang, Xiaobao Wei, Wenzhao Zheng, Pengju An, Ming Lu, Wei Zhan, Masayoshi Tomizuka, Kurt Keutzer, and Shanghang Zhang.
\newblock S3gaussian: Self-supervised street gaussians for autonomous driving.
\newblock \emph{arXiv preprint arXiv:2405.20323}, 2024.

\bibitem[Huang et~al.(2022)Huang, Du, Yang, Zhou, Zhang, and Chen]{prediction1}
Yanjun Huang, Jiatong Du, Ziru Yang, Zewei Zhou, Lin Zhang, and Hong Chen.
\newblock A survey on trajectory-prediction methods for autonomous driving.
\newblock \emph{IEEE Transactions on Intelligent Vehicles}, 7\penalty0 (3):\penalty0 652--674, 2022.

\bibitem[Huang et~al.(2025)Huang, Wang, Zhang, Zhang, Li, and Ji]{huang2025givepose}
Ziqin Huang, Gu Wang, Chenyangguang Zhang, Ruida Zhang, Xiu Li, and Xiangyang Ji.
\newblock Givepose: Gradual intra-class variation elimination for rgb-based category-level object pose estimation.
\newblock In \emph{Proceedings of the Computer Vision and Pattern Recognition Conference}, pages 22055--22066, 2025.

\bibitem[Jiang et~al.(2023)Jiang, Chen, Xu, Liao, Chen, Zhou, Zhang, Liu, Huang, and Wang]{end-to-end3}
Bo Jiang, Shaoyu Chen, Qing Xu, Bencheng Liao, Jiajie Chen, Helong Zhou, Qian Zhang, Wenyu Liu, Chang Huang, and Xinggang Wang.
\newblock Vad: Vectorized scene representation for efficient autonomous driving.
\newblock In \emph{Proceedings of the IEEE/CVF International Conference on Computer Vision}, pages 8340--8350, 2023.

\bibitem[Kalake et~al.(2021)Kalake, Wan, and Hou]{MOTreview2021_2}
Lesole Kalake, Wanggen Wan, and Li Hou.
\newblock Analysis based on recent deep learning approaches applied in real-time multi-object tracking: a review.
\newblock \emph{IEEE Access}, 9:\penalty0 32650--32671, 2021.

\bibitem[Kerbl et~al.(2023)Kerbl, Kopanas, Leimk{\"u}hler, and Drettakis]{3dgs}
Bernhard Kerbl, Georgios Kopanas, Thomas Leimk{\"u}hler, and George Drettakis.
\newblock 3d gaussian splatting for real-time radiance field rendering.
\newblock \emph{ACM Trans. Graph.}, 42\penalty0 (4):\penalty0 139--1, 2023.

\bibitem[Khan et~al.(2024)Khan, Fazlali, Sharma, Cao, Bai, Ren, and Liu]{khan2024autosplat}
Mustafa Khan, Hamidreza Fazlali, Dhruv Sharma, Tongtong Cao, Dongfeng Bai, Yuan Ren, and Bingbing Liu.
\newblock Autosplat: Constrained gaussian splatting for autonomous driving scene reconstruction.
\newblock \emph{arXiv preprint arXiv:2407.02598}, 2024.

\bibitem[Kingma(2014)]{kingma2014adam}
Diederik~P Kingma.
\newblock Adam: A method for stochastic optimization.
\newblock \emph{arXiv preprint arXiv:1412.6980}, 2014.

\bibitem[Kirillov et~al.(2023)Kirillov, Mintun, Ravi, Mao, Rolland, Gustafson, Xiao, Whitehead, Berg, Lo, et~al.]{kirillov2023segmentanything}
Alexander Kirillov, Eric Mintun, Nikhila Ravi, Hanzi Mao, Chloe Rolland, Laura Gustafson, Tete Xiao, Spencer Whitehead, Alexander~C Berg, Wan-Yen Lo, et~al.
\newblock Segment anything.
\newblock In \emph{Proceedings of the IEEE/CVF International Conference on Computer Vision}, pages 4015--4026, 2023.

\bibitem[Kuznetsova et~al.(2020)Kuznetsova, Rom, Alldrin, Uijlings, Krasin, Pont-Tuset, Kamali, Popov, Malloci, Kolesnikov, et~al.]{kuznet2020openimages}
Alina Kuznetsova, Hassan Rom, Neil Alldrin, Jasper Uijlings, Ivan Krasin, Jordi Pont-Tuset, Shahab Kamali, Stefan Popov, Matteo Malloci, Alexander Kolesnikov, et~al.
\newblock The open images dataset v4: Unified image classification, object detection, and visual relationship detection at scale.
\newblock \emph{International journal of computer vision}, 128\penalty0 (7):\penalty0 1956--1981, 2020.

\bibitem[Li et~al.(2023)Li, Luo, and Yang]{li2023pillarnext}
Jinyu Li, Chenxu Luo, and Xiaodong Yang.
\newblock Pillarnext: Rethinking network designs for 3d object detection in lidar point clouds.
\newblock In \emph{Proceedings of the IEEE/CVF Conference on Computer Vision and Pattern Recognition}, pages 17567--17576, 2023.

\bibitem[Li et~al.(2019)Li, Pan, Zhang, Ren, Ma, Fang, Yan, Geng, Huang, Gong, et~al.]{li2019aads}
Wei Li, CW Pan, Rong Zhang, JP Ren, YX Ma, Jin Fang, FL Yan, QC Geng, XY Huang, HJ Gong, et~al.
\newblock Aads: Augmented autonomous driving simulation using data-driven algorithms.
\newblock \emph{Science robotics}, 4\penalty0 (28):\penalty0 eaaw0863, 2019.

\bibitem[Li et~al.(2024{\natexlab{a}})Li, Lyu, Di, Zhai, Lee, and Tombari]{li2024geogaussian}
Yanyan Li, Chenyu Lyu, Yan Di, Guangyao Zhai, Gim~Hee Lee, and Federico Tombari.
\newblock Geogaussian: Geometry-aware gaussian splatting for scene rendering.
\newblock In \emph{European Conference on Computer Vision}, pages 441--457. Springer, 2024{\natexlab{a}}.

\bibitem[Li et~al.(2017)Li, Ji, Yang, Ye, and Yu]{li2017robust}
Zhong Li, Yu Ji, Wei Yang, Jinwei Ye, and Jingyi Yu.
\newblock Robust 3d human motion reconstruction via dynamic template construction.
\newblock In \emph{2017 International Conference on 3D Vision (3DV)}, pages 496--505. IEEE, 2017.

\bibitem[Li et~al.(2024{\natexlab{b}})Li, Yu, Lan, Li, Kautz, Lu, and Alvarez]{open-loop-eval2}
Zhiqi Li, Zhiding Yu, Shiyi Lan, Jiahan Li, Jan Kautz, Tong Lu, and Jose~M Alvarez.
\newblock Is ego status all you need for open-loop end-to-end autonomous driving?
\newblock In \emph{Proceedings of the IEEE/CVF Conference on Computer Vision and Pattern Recognition}, pages 14864--14873, 2024{\natexlab{b}}.

\bibitem[Liang et~al.(2020)Liang, Yang, Zeng, Chen, Hu, Casas, and Urtasun]{prediction2}
Ming Liang, Bin Yang, Wenyuan Zeng, Yun Chen, Rui Hu, Sergio Casas, and Raquel Urtasun.
\newblock Pnpnet: End-to-end perception and prediction with tracking in the loop.
\newblock In \emph{Proceedings of the IEEE/CVF Conference on Computer Vision and Pattern Recognition}, pages 11553--11562, 2020.

\bibitem[Liang et~al.(2023)Liang, Khan, Li, Nguyen-Phuoc, Lanman, Tompkin, and Xiao]{liang2023gaufre}
Yiqing Liang, Numair Khan, Zhengqin Li, Thu Nguyen-Phuoc, Douglas Lanman, James Tompkin, and Lei Xiao.
\newblock Gaufre: Gaussian deformation fields for real-time dynamic novel view synthesis.
\newblock \emph{arXiv preprint arXiv:2312.11458}, 2023.

\bibitem[Liao et~al.(2022)Liao, Xie, and Geiger]{liao2022kitti}
Yiyi Liao, Jun Xie, and Andreas Geiger.
\newblock Kitti-360: A novel dataset and benchmarks for urban scene understanding in 2d and 3d.
\newblock \emph{IEEE Transactions on Pattern Analysis and Machine Intelligence}, 45\penalty0 (3):\penalty0 3292--3310, 2022.

\bibitem[Lin et~al.(2014)Lin, Maire, Belongie, Hays, Perona, Ramanan, Doll{\'a}r, and Zitnick]{lin2014coco}
Tsung-Yi Lin, Michael Maire, Serge Belongie, James Hays, Pietro Perona, Deva Ramanan, Piotr Doll{\'a}r, and C~Lawrence Zitnick.
\newblock Microsoft coco: Common objects in context.
\newblock In \emph{Computer Vision--ECCV 2014: 13th European Conference, Zurich, Switzerland, September 6-12, 2014, Proceedings, Part V 13}, pages 740--755. Springer, 2014.

\bibitem[Liu et~al.(2022)Liu, Wang, Li, and Ji]{liu_2022_catre}
Xingyu Liu, Gu Wang, Yi Li, and Xiangyang Ji.
\newblock {CATRE:} iterative point clouds alignment for category-level object pose refinement.
\newblock In \emph{European Conference on Computer Vision}, 2022.

\bibitem[Liu et~al.(2024)Liu, Zhang, Wang, Zhang, and Ji]{liu2024rasim}
Xingyu Liu, Chenyangguang Zhang, Gu Wang, Ruida Zhang, and Xiangyang Ji.
\newblock Rasim: A range-aware high-fidelity rgb-d data simulation pipeline for real-world applications.
\newblock In \emph{2024 IEEE international conference on robotics and automation (ICRA)}, pages 17057--17064. IEEE, 2024.

\bibitem[Liu et~al.(2025{\natexlab{a}})Liu, Li, Li, Wang, Zhang, Huang, and Ji]{liu2024gfreedet}
Xingyu Liu, Yingyue Li, Chengxi Li, Gu Wang, Chenyangguang Zhang, Ziqin Huang, and Xiangyang Ji.
\newblock Gfreedet: Exploiting gaussian splatting and foundation models for model-free unseen object detection in the bop challenge 2024.
\newblock \emph{IEEE/CVF Conference on Computer Vision and Pattern Recognition Workshop}, 2025{\natexlab{a}}.

\bibitem[Liu et~al.(2025{\natexlab{b}})Liu, Wang, Zhang, Zhang, Tombari, and Ji]{liu_2025_unopose}
Xingyu Liu, Gu Wang, Ruida Zhang, Chenyangguang Zhang, Federico Tombari, and Xiangyang Ji.
\newblock {UNOPose}: Unseen object pose estimation with an unposed rgb-d reference image.
\newblock In \emph{Proceedings of the IEEE/CVF Conference on Computer Vision and Pattern Recognition}, 2025{\natexlab{b}}.

\bibitem[Liu et~al.(2025{\natexlab{c}})Liu, Zhang, Zhang, Wang, Tang, Li, and Ji]{liu2025gdrnpp}
Xingyu Liu, Ruida Zhang, Chenyangguang Zhang, Gu Wang, Jiwen Tang, Zhigang Li, and Xiangyang Ji.
\newblock Gdrnpp: A geometry-guided and fully learning-based object pose estimator.
\newblock \emph{IEEE Transactions on Pattern Analysis and Machine Intelligence}, 2025{\natexlab{c}}.

\bibitem[Liu et~al.(2023)Liu, Tang, Amini, Yang, Mao, Rus, and Han]{liu2023bevfusion}
Zhijian Liu, Haotian Tang, Alexander Amini, Xinyu Yang, Huizi Mao, Daniela~L Rus, and Song Han.
\newblock Bevfusion: Multi-task multi-sensor fusion with unified bird's-eye view representation.
\newblock In \emph{2023 IEEE international conference on robotics and automation (ICRA)}, pages 2774--2781. IEEE, 2023.

\bibitem[Luiten et~al.(2023)Luiten, Kopanas, Leibe, and Ramanan]{luiten2023dynamic3dgs}
Jonathon Luiten, Georgios Kopanas, Bastian Leibe, and Deva Ramanan.
\newblock Dynamic 3d gaussians: Tracking by persistent dynamic view synthesis.
\newblock \emph{arXiv preprint arXiv:2308.09713}, 2023.

\bibitem[Luo et~al.(2021)Luo, Xing, Milan, Zhang, Liu, and Kim]{MOTreview21}
Wenhan Luo, Junliang Xing, Anton Milan, Xiaoqin Zhang, Wei Liu, and Tae-Kyun Kim.
\newblock Multiple object tracking: A literature review.
\newblock \emph{Artificial intelligence}, 293:\penalty0 103448, 2021.

\bibitem[Mao et~al.(2021)Mao, Niu, Jiang, Liang, Chen, Liang, Li, Ye, Zhang, Li, et~al.]{mao2021oncedataset}
Jiageng Mao, Minzhe Niu, Chenhan Jiang, Hanxue Liang, Jingheng Chen, Xiaodan Liang, Yamin Li, Chaoqiang Ye, Wei Zhang, Zhenguo Li, et~al.
\newblock One million scenes for autonomous driving: Once dataset.
\newblock \emph{arXiv preprint arXiv:2106.11037}, 2021.

\bibitem[Mao et~al.(2023)Mao, Shi, Wang, and Li]{3Ddet_survey23}
Jiageng Mao, Shaoshuai Shi, Xiaogang Wang, and Hongsheng Li.
\newblock 3d object detection for autonomous driving: A comprehensive survey.
\newblock \emph{International Journal of Computer Vision}, 131\penalty0 (8):\penalty0 1909--1963, 2023.

\bibitem[Mildenhall et~al.(2021)Mildenhall, Srinivasan, Tancik, Barron, Ramamoorthi, and Ng]{mildenhall2021nerf}
Ben Mildenhall, Pratul~P Srinivasan, Matthew Tancik, Jonathan~T Barron, Ravi Ramamoorthi, and Ren Ng.
\newblock Nerf: Representing scenes as neural radiance fields for view synthesis.
\newblock \emph{Communications of the ACM}, 65\penalty0 (1):\penalty0 99--106, 2021.

\bibitem[Minderer et~al.(2022)Minderer, Gritsenko, Stone, Neumann, Weissenborn, Dosovitskiy, Mahendran, Arnab, Dehghani, Shen, et~al.]{owl-vit}
M Minderer, A Gritsenko, A Stone, M Neumann, D Weissenborn, A Dosovitskiy, A Mahendran, A Arnab, M Dehghani, Z Shen, et~al.
\newblock Simple open-vocabulary object detection with vision transformers. arxiv 2022.
\newblock \emph{arXiv preprint arXiv:2205.06230}, 2, 2022.

\bibitem[M{\"u}ller et~al.(2022)M{\"u}ller, Evans, Schied, and Keller]{muller2022instant_neural}
Thomas M{\"u}ller, Alex Evans, Christoph Schied, and Alexander Keller.
\newblock Instant neural graphics primitives with a multiresolution hash encoding.
\newblock \emph{ACM transactions on graphics (TOG)}, 41\penalty0 (4):\penalty0 1--15, 2022.

\bibitem[Oquab et~al.(2023)Oquab, Darcet, Moutakanni, Vo, Szafraniec, Khalidov, Fernandez, Haziza, Massa, El-Nouby, et~al.]{oquab2023dinov2}
Maxime Oquab, Timoth{\'e}e Darcet, Th{\'e}o Moutakanni, Huy Vo, Marc Szafraniec, Vasil Khalidov, Pierre Fernandez, Daniel Haziza, Francisco Massa, Alaaeldin El-Nouby, et~al.
\newblock Dinov2: Learning robust visual features without supervision.
\newblock \emph{arXiv preprint arXiv:2304.07193}, 2023.

\bibitem[Ost et~al.(2021)Ost, Mannan, Thuerey, Knodt, and Heide]{ost2021nsg}
Julian Ost, Fahim Mannan, Nils Thuerey, Julian Knodt, and Felix Heide.
\newblock Neural scene graphs for dynamic scenes.
\newblock In \emph{Proceedings of the IEEE/CVF Conference on Computer Vision and Pattern Recognition}, pages 2856--2865, 2021.

\bibitem[Pang et~al.(2022)Pang, Li, and Wang]{pang2022simpletrack}
Ziqi Pang, Zhichao Li, and Naiyan Wang.
\newblock Simpletrack: Understanding and rethinking 3d multi-object tracking.
\newblock In \emph{European Conference on Computer Vision}, pages 680--696. Springer, 2022.

\bibitem[Peng et~al.(2025)Peng, Zhang, Wang, Xu, Xie, Zheng, Keutzer, Tomizuka, and Zhan]{peng2025desire}
Chensheng Peng, Chengwei Zhang, Yixiao Wang, Chenfeng Xu, Yichen Xie, Wenzhao Zheng, Kurt Keutzer, Masayoshi Tomizuka, and Wei Zhan.
\newblock Desire-gs: 4d street gaussians for static-dynamic decomposition and surface reconstruction for urban driving scenes.
\newblock In \emph{Proceedings of the Computer Vision and Pattern Recognition Conference}, pages 6782--6791, 2025.

\bibitem[Radford et~al.(2021)Radford, Kim, Hallacy, Ramesh, Goh, Agarwal, Sastry, Askell, Mishkin, Clark, et~al.]{radford2021clip}
Alec Radford, Jong~Wook Kim, Chris Hallacy, Aditya Ramesh, Gabriel Goh, Sandhini Agarwal, Girish Sastry, Amanda Askell, Pamela Mishkin, Jack Clark, et~al.
\newblock Learning transferable visual models from natural language supervision.
\newblock In \emph{International conference on machine learning}, pages 8748--8763. PMLR, 2021.

\bibitem[Sanchez et~al.(2023)Sanchez, Deschaud, and Goulette]{sanchez2023dgin3dsegment}
Jules Sanchez, Jean-Emmanuel Deschaud, and Fran{\c{c}}ois Goulette.
\newblock Domain generalization of 3d semantic segmentation in autonomous driving.
\newblock In \emph{Proceedings of the IEEE/CVF International Conference on Computer Vision}, pages 18077--18087, 2023.

\bibitem[{Sara Fridovich-Keil and Giacomo Meanti} et~al.(2023){Sara Fridovich-Keil and Giacomo Meanti}, Warburg, Recht, and Kanazawa]{kplanes_2023}
{Sara Fridovich-Keil and Giacomo Meanti}, Frederik~Rahbæk Warburg, Benjamin Recht, and Angjoo Kanazawa.
\newblock K-planes: Explicit radiance fields in space, time, and appearance.
\newblock In \emph{CVPR}, 2023.

\bibitem[Shah et~al.(2018)Shah, Dey, Lovett, and Kapoor]{shah2018airsim}
Shital Shah, Debadeepta Dey, Chris Lovett, and Ashish Kapoor.
\newblock Airsim: High-fidelity visual and physical simulation for autonomous vehicles.
\newblock In \emph{Field and Service Robotics: Results of the 11th International Conference}, pages 621--635. Springer, 2018.

\bibitem[Song et~al.(2022)Song, Kong, Lee, Kwak, and Lee]{song2022towards_effi}
Yeji Song, Chaerin Kong, Seoyoung Lee, Nojun Kwak, and Joonseok Lee.
\newblock Towards efficient neural scene graphs by learning consistency fields.
\newblock \emph{arXiv preprint arXiv:2210.04127}, 2022.

\bibitem[Soum-Fontez et~al.(2023)Soum-Fontez, Deschaud, and Goulette]{soum2023mdt3d}
Louis Soum-Fontez, Jean-Emmanuel Deschaud, and Fran{\c{c}}ois Goulette.
\newblock Mdt3d: Multi-dataset training for lidar 3d object detection generalization.
\newblock In \emph{2023 IEEE/RSJ International Conference on Intelligent Robots and Systems (IROS)}, pages 5765--5772. IEEE, 2023.

\bibitem[Stearns et~al.(2024)Stearns, Harley, Uy, Dubost, Tombari, Wetzstein, and Guibas]{stearns2024marble}
Colton Stearns, Adam Harley, Mikaela Uy, Florian Dubost, Federico Tombari, Gordon Wetzstein, and Leonidas Guibas.
\newblock Dynamic gaussian marbles for novel view synthesis of casual monocular videos.
\newblock \emph{arXiv preprint arXiv:2406.18717}, 2024.

\bibitem[Su et~al.(2020)Su, Xu, Zheng, Yu, Liu, and Fang]{su2020robustfusion}
Zhuo Su, Lan Xu, Zerong Zheng, Tao Yu, Yebin Liu, and Lu Fang.
\newblock Robustfusion: Human volumetric capture with data-driven visual cues using a rgbd camera.
\newblock In \emph{Computer Vision--ECCV 2020: 16th European Conference, Glasgow, UK, August 23--28, 2020, Proceedings, Part IV 16}, pages 246--264. Springer, 2020.

\bibitem[Sun et~al.(2020)Sun, Kretzschmar, Dotiwalla, Chouard, Patnaik, Tsui, Guo, Zhou, Chai, Caine, et~al.]{sun2020waymo}
Pei Sun, Henrik Kretzschmar, Xerxes Dotiwalla, Aurelien Chouard, Vijaysai Patnaik, Paul Tsui, James Guo, Yin Zhou, Yuning Chai, Benjamin Caine, et~al.
\newblock Scalability in perception for autonomous driving: Waymo open dataset.
\newblock In \emph{Proceedings of the IEEE/CVF conference on computer vision and pattern recognition}, pages 2446--2454, 2020.

\bibitem[Sun et~al.(2024)Sun, Zhuang, Jiang, Liu, Xie, and Chandraker]{sun2024lidarf}
Shanlin Sun, Bingbing Zhuang, Ziyu Jiang, Buyu Liu, Xiaohui Xie, and Manmohan Chandraker.
\newblock Lidarf: Delving into lidar for neural radiance field on street scenes.
\newblock In \emph{Proceedings of the IEEE/CVF Conference on Computer Vision and Pattern Recognition}, pages 19563--19572, 2024.

\bibitem[Turki et~al.(2023)Turki, Zhang, Ferroni, and Ramanan]{turki2023suds}
Haithem Turki, Jason~Y Zhang, Francesco Ferroni, and Deva Ramanan.
\newblock Suds: Scalable urban dynamic scenes.
\newblock In \emph{Proceedings of the IEEE/CVF Conference on Computer Vision and Pattern Recognition}, pages 12375--12385, 2023.

\bibitem[Wang et~al.(2020)Wang, Chen, You, Li, Hariharan, Campbell, Weinberger, and Chao]{train_in_Germany}
Yan Wang, Xiangyu Chen, Yurong You, Li~Erran Li, Bharath Hariharan, Mark Campbell, Kilian~Q Weinberger, and Wei-Lun Chao.
\newblock Train in germany, test in the usa: Making 3d object detectors generalize.
\newblock In \emph{Proceedings of the IEEE/CVF Conference on Computer Vision and Pattern Recognition}, pages 11713--11723, 2020.

\bibitem[Wang et~al.(2024)Wang, Wang, Li, and Liu]{3Ddet_survey24}
Yu Wang, Shaohua Wang, Yicheng Li, and Mingchun Liu.
\newblock A comprehensive review of 3d object detection in autonomous driving: Technological advances and future directions.
\newblock \emph{arXiv preprint arXiv:2408.16530}, 2024.

\bibitem[Wu et~al.(2024{\natexlab{a}})Wu, Yi, Fang, Xie, Zhang, Wei, Liu, Tian, and Wang]{4dgs}
Guanjun Wu, Taoran Yi, Jiemin Fang, Lingxi Xie, Xiaopeng Zhang, Wei Wei, Wenyu Liu, Qi Tian, and Xinggang Wang.
\newblock 4d gaussian splatting for real-time dynamic scene rendering.
\newblock In \emph{Proceedings of the IEEE/CVF Conference on Computer Vision and Pattern Recognition (CVPR)}, pages 20310--20320, 2024{\natexlab{a}}.

\bibitem[Wu et~al.(2021)Wu, Han, Wen, Li, and Wang]{wu2021castrack}
Hai Wu, Wenkai Han, Chenglu Wen, Xin Li, and Cheng Wang.
\newblock 3d multi-object tracking in point clouds based on prediction confidence-guided data association.
\newblock \emph{IEEE Transactions on Intelligent Transportation Systems}, 23\penalty0 (6):\penalty0 5668--5677, 2021.

\bibitem[Wu et~al.(2022)Wu, Deng, Wen, Li, Wang, and Li]{wu2022casa}
Hai Wu, Jinhao Deng, Chenglu Wen, Xin Li, Cheng Wang, and Jonathan Li.
\newblock Casa: A cascade attention network for 3-d object detection from lidar point clouds.
\newblock \emph{IEEE Transactions on Geoscience and Remote Sensing}, 60:\penalty0 1--11, 2022.

\bibitem[Wu et~al.(2024{\natexlab{b}})Wu, Jiang, Liu, Yuan, Bai, and Bai]{wu2024glee}
Junfeng Wu, Yi Jiang, Qihao Liu, Zehuan Yuan, Xiang Bai, and Song Bai.
\newblock General object foundation model for images and videos at scale.
\newblock In \emph{Proceedings of the IEEE/CVF Conference on Computer Vision and Pattern Recognition}, pages 3783--3795, 2024{\natexlab{b}}.

\bibitem[Wu et~al.(2023)Wu, Liu, Luo, Zhong, Chen, Xiao, Hou, Lou, Chen, Yang, et~al.]{wu2023mars}
Zirui Wu, Tianyu Liu, Liyi Luo, Zhide Zhong, Jianteng Chen, Hongmin Xiao, Chao Hou, Haozhe Lou, Yuantao Chen, Runyi Yang, et~al.
\newblock Mars: An instance-aware, modular and realistic simulator for autonomous driving.
\newblock In \emph{CAAI International Conference on Artificial Intelligence}, pages 3--15. Springer, 2023.

\bibitem[Xiao et~al.(2021)Xiao, Shao, Hao, Zhang, Chai, Jiao, Li, Wu, Sun, Jiang, et~al.]{xiao2021pandaset}
Pengchuan Xiao, Zhenlei Shao, Steven Hao, Zishuo Zhang, Xiaolin Chai, Judy Jiao, Zesong Li, Jian Wu, Kai Sun, Kun Jiang, et~al.
\newblock Pandaset: Advanced sensor suite dataset for autonomous driving.
\newblock In \emph{2021 IEEE International Intelligent Transportation Systems Conference (ITSC)}, pages 3095--3101. IEEE, 2021.

\bibitem[Xie et~al.(2023)Xie, Zhang, Li, Zhang, and Zhang]{xie2023s_nerf}
Ziyang Xie, Junge Zhang, Wenye Li, Feihu Zhang, and Li Zhang.
\newblock S-nerf: Neural radiance fields for street views.
\newblock \emph{arXiv preprint arXiv:2303.00749}, 2023.

\bibitem[Xu et~al.(2022{\natexlab{a}})Xu, Zhang, Zhang, and Tao]{xu2022vitpose}
Yufei Xu, Jing Zhang, Qiming Zhang, and Dacheng Tao.
\newblock Vi{TP}ose: Simple vision transformer baselines for human pose estimation.
\newblock In \emph{Advances in Neural Information Processing Systems}, 2022{\natexlab{a}}.

\bibitem[Xu et~al.(2022{\natexlab{b}})Xu, Zhang, Zhang, and Tao]{xu2022vitpose+}
Yufei Xu, Jing Zhang, Qiming Zhang, and Dacheng Tao.
\newblock Vitpose+: Vision transformer foundation model for generic body pose estimation.
\newblock \emph{arXiv preprint arXiv:2212.04246}, 2022{\natexlab{b}}.

\bibitem[Yan et~al.(2023)Yan, Jiang, Wu, Wang, Luo, Yuan, and Lu]{yan2023uninext}
Bin Yan, Yi Jiang, Jiannan Wu, Dong Wang, Ping Luo, Zehuan Yuan, and Huchuan Lu.
\newblock Universal instance perception as object discovery and retrieval.
\newblock In \emph{Proceedings of the IEEE/CVF Conference on Computer Vision and Pattern Recognition}, pages 15325--15336, 2023.

\bibitem[Yan et~al.(2024)Yan, Lin, Zhou, Wang, Sun, Zhan, Lang, Zhou, and Peng]{yan2024streetgaussian}
Yunzhi Yan, Haotong Lin, Chenxu Zhou, Weijie Wang, Haiyang Sun, Kun Zhan, Xianpeng Lang, Xiaowei Zhou, and Sida Peng.
\newblock Street gaussians: Modeling dynamic urban scenes with gaussian splatting.
\newblock In \emph{ECCV}, 2024.

\bibitem[Yang et~al.(2023{\natexlab{a}})Yang, Ivanovic, Litany, Weng, Kim, Li, Che, Xu, Fidler, Pavone, et~al.]{yang2023emernerf}
Jiawei Yang, Boris Ivanovic, Or Litany, Xinshuo Weng, Seung~Wook Kim, Boyi Li, Tong Che, Danfei Xu, Sanja Fidler, Marco Pavone, et~al.
\newblock Emernerf: Emergent spatial-temporal scene decomposition via self-supervision.
\newblock \emph{arXiv preprint arXiv:2311.02077}, 2023{\natexlab{a}}.

\bibitem[Yang et~al.(2023{\natexlab{b}})Yang, Chen, Wang, Manivasagam, Ma, Yang, and Urtasun]{yang2023unisim}
Ze Yang, Yun Chen, Jingkang Wang, Sivabalan Manivasagam, Wei-Chiu Ma, Anqi~Joyce Yang, and Raquel Urtasun.
\newblock Unisim: A neural closed-loop sensor simulator.
\newblock In \emph{Proceedings of the IEEE/CVF Conference on Computer Vision and Pattern Recognition}, pages 1389--1399, 2023{\natexlab{b}}.

\bibitem[Yang et~al.(2023{\natexlab{c}})Yang, Yang, Pan, and Zhang]{yang2023realtime4d}
Zeyu Yang, Hongye Yang, Zijie Pan, and Li Zhang.
\newblock Real-time photorealistic dynamic scene representation and rendering with 4d gaussian splatting.
\newblock \emph{arXiv preprint arXiv:2310.10642}, 2023{\natexlab{c}}.

\bibitem[Yang et~al.(2024)Yang, Gao, Zhou, Jiao, Zhang, and Jin]{yang2024deformable3dgs}
Ziyi Yang, Xinyu Gao, Wen Zhou, Shaohui Jiao, Yuqing Zhang, and Xiaogang Jin.
\newblock Deformable 3d gaussians for high-fidelity monocular dynamic scene reconstruction.
\newblock In \emph{Proceedings of the IEEE/CVF Conference on Computer Vision and Pattern Recognition}, pages 20331--20341, 2024.

\bibitem[Yin et~al.(2021)Yin, Zhou, and Krahenbuhl]{yin2021centerpoint}
Tianwei Yin, Xingyi Zhou, and Philipp Krahenbuhl.
\newblock Center-based 3d object detection and tracking.
\newblock In \emph{Proceedings of the IEEE/CVF conference on computer vision and pattern recognition}, pages 11784--11793, 2021.

\bibitem[Yu et~al.(2018)Yu, Xian, Chen, Liu, Liao, Madhavan, Darrell, et~al.]{yu2018bdd100k}
Fisher Yu, Wenqi Xian, Yingying Chen, Fangchen Liu, Mike Liao, Vashisht Madhavan, Trevor Darrell, et~al.
\newblock Bdd100k: A diverse driving video database with scalable annotation tooling.
\newblock \emph{arXiv preprint arXiv:1805.04687}, 2\penalty0 (5):\penalty0 6, 2018.

\bibitem[Zhai et~al.(2023)Zhai, Feng, Du, Mao, Liu, Tan, Zhang, Ye, and Wang]{open-loop-eval1}
Jiang-Tian Zhai, Ze Feng, Jinhao Du, Yongqiang Mao, Jiang-Jiang Liu, Zichang Tan, Yifu Zhang, Xiaoqing Ye, and Jingdong Wang.
\newblock Rethinking the open-loop evaluation of end-to-end autonomous driving in nuscenes.
\newblock \emph{arXiv preprint arXiv:2305.10430}, 2023.

\bibitem[Zhang et~al.(2023{\natexlab{a}})Zhang, Yuan, Shi, Chen, Li, and Qiao]{zhang2023uni3d}
Bo Zhang, Jiakang Yuan, Botian Shi, Tao Chen, Yikang Li, and Yu Qiao.
\newblock Uni3d: A unified baseline for multi-dataset 3d object detection.
\newblock In \emph{Proceedings of the IEEE/CVF Conference on Computer Vision and Pattern Recognition}, pages 9253--9262, 2023{\natexlab{a}}.

\bibitem[Zhang et~al.(2023{\natexlab{b}})Zhang, Li, He, and Lin]{3DMOT_review23}
Peng Zhang, Xin Li, Liang He, and Xin Lin.
\newblock 3d multiple object tracking on autonomous driving: A literature review.
\newblock \emph{arXiv preprint arXiv:2309.15411}, 2023{\natexlab{b}}.

\bibitem[Zhang et~al.(2022{\natexlab{a}})Zhang, Di, Lou, Manhardt, Tombari, and Ji]{zhang2022rbp}
Ruida Zhang, Yan Di, Zhiqiang Lou, Fabian Manhardt, Federico Tombari, and Xiangyang Ji.
\newblock Rbp-pose: Residual bounding box projection for category-level pose estimation.
\newblock In \emph{European conference on computer vision}, pages 655--672. Springer, 2022{\natexlab{a}}.

\bibitem[Zhang et~al.(2022{\natexlab{b}})Zhang, Di, Manhardt, Tombari, and Ji]{zhang2022ssp}
Ruida Zhang, Yan Di, Fabian Manhardt, Federico Tombari, and Xiangyang Ji.
\newblock Ssp-pose: Symmetry-aware shape prior deformation for direct category-level object pose estimation.
\newblock In \emph{2022 IEEE/RSJ International Conference on Intelligent Robots and Systems (IROS)}, pages 7452--7459. IEEE, 2022{\natexlab{b}}.

\bibitem[Zhang et~al.(2024{\natexlab{a}})Zhang, Huang, Wang, Zhang, Di, Zuo, Tang, and Ji]{zhang2024lapose}
Ruida Zhang, Ziqin Huang, Gu Wang, Chenyangguang Zhang, Yan Di, Xingxing Zuo, Jiwen Tang, and Xiangyang Ji.
\newblock Lapose: Laplacian mixture shape modeling for rgb-based category-level object pose estimation.
\newblock In \emph{European Conference on Computer Vision}, pages 467--484. Springer, 2024{\natexlab{a}}.

\bibitem[Zhang et~al.(2024{\natexlab{b}})Zhang, Zhang, Di, Manhardt, Liu, Tombari, and Ji]{zhang2024kpred}
Ruida Zhang, Chenyangguang Zhang, Yan Di, Fabian Manhardt, Xingyu Liu, Federico Tombari, and Xiangyang Ji.
\newblock Kp-red: Exploiting semantic keypoints for joint 3d shape retrieval and deformation.
\newblock In \emph{Proceedings of the IEEE/CVF Conference on Computer Vision and Pattern Recognition}, pages 20540--20550, 2024{\natexlab{b}}.

\bibitem[Zhang et~al.(2022{\natexlab{c}})Zhang, Xu, Xu, Liu, Cui, Wan, Sun, and Li]{zhang2022det_generaliz}
Xingxuan Zhang, Zekai Xu, Renzhe Xu, Jiashuo Liu, Peng Cui, Weitao Wan, Chong Sun, and Chen Li.
\newblock Towards domain generalization in object detection.
\newblock \emph{arXiv preprint arXiv:2203.14387}, 2022{\natexlab{c}}.

\bibitem[Zhang et~al.(2021)Zhang, Wang, Wang, Zeng, and Liu]{zhang2021fairmot}
Yifu Zhang, Chunyu Wang, Xinggang Wang, Wenjun Zeng, and Wenyu Liu.
\newblock Fairmot: On the fairness of detection and re-identification in multiple object tracking.
\newblock \emph{International journal of computer vision}, 129:\penalty0 3069--3087, 2021.

\bibitem[Zhou et~al.(2024{\natexlab{a}})Zhou, Shao, Xu, Bai, Qiu, Liu, Wang, Geiger, and Liao]{zhou2024hugs}
Hongyu Zhou, Jiahao Shao, Lu Xu, Dongfeng Bai, Weichao Qiu, Bingbing Liu, Yue Wang, Andreas Geiger, and Yiyi Liao.
\newblock Hugs: Holistic urban 3d scene understanding via gaussian splatting.
\newblock In \emph{Proceedings of the IEEE/CVF Conference on Computer Vision and Pattern Recognition}, pages 21336--21345, 2024{\natexlab{a}}.

\bibitem[Zhou et~al.(2024{\natexlab{b}})Zhou, Lin, Shan, Wang, Sun, and Yang]{zhou2024drivinggaussian}
Xiaoyu Zhou, Zhiwei Lin, Xiaojun Shan, Yongtao Wang, Deqing Sun, and Ming-Hsuan Yang.
\newblock Drivinggaussian: Composite gaussian splatting for surrounding dynamic autonomous driving scenes.
\newblock In \emph{Proceedings of the IEEE/CVF Conference on Computer Vision and Pattern Recognition}, pages 21634--21643, 2024{\natexlab{b}}.

\end{thebibliography}
}

\clearpage
\setcounter{page}{1}
\maketitlesupplementary

\section{Evaluation on Static Scenes in Waymo-NOTR Dataset}
We evaluate our method on the static32 subset of the Waymo-NOTR dataset \cite{yang2023emernerf, sun2020waymo}, following the experimental setup of EmerNeRF \cite{yang2023emernerf} for novel view synthesis. 
As shown in Table \ref{tab:static}, while our primary focus is on handling dynamic objects, our method also demonstrates robust performance in static scenes.

\renewcommand{\arraystretch}{1.25}
\begin{table}[!ht]
    \centering
    \begin{tabular}{c|ccc}
    \hline
        \small Method & \small PSNR↑ & \small SSIM↑ & \small LPIPS↓ \\ \hline
        \small 3DGS \cite{3dgs}  & 26.82 & 0.836 & 0.134 \\ 
        \small EmerNeRF \cite{yang2023emernerf} & {\cellcolor{red!15}28.89} & 0.814 & 0.212  \\ 
        \small S3Gaussian \cite{s3gaussian} & {27.05} & {0.825} & \cellcolor{cyan!15}0.142  \\ 
        \small MARS \cite{wu2023mars}& {27.63} & {\cellcolor{cyan!15}0.848} & 0.193 \\ 
        \small Ours  & {\cellcolor{cyan!15}28.72} & {\cellcolor{red!15}0.857} & {\cellcolor{red!15}0.092}
        \\ \hline
    \end{tabular}
    \caption{Comparison with state-of-the-art methods on the static32 subset of Waymo-NOTR dataset. StreetGS represents Street Gaussian \cite{yan2024streetgaussian}. The \colorbox{red!15}{best} and the \colorbox{cyan!15}{second best} results are denoted by pink and blue. 
    }
    \label{tab:static}
\end{table}

\section{Runtime Analysis}
As shown in Table \ref{tab:speed}, we evaluate the inference speed of our method and several state-of-the-art methods at a resolution of $960 \times 640$ on the same device.
\begin{table}[!ht]
    \centering
    \begin{tabular}{c|ccc|c}
    \hline
    Method &  3DGS \cite{3dgs} & S3G \cite{s3gaussian} &  SG \cite{yan2024streetgaussian} & Ours \\ \hline
        Speed (FPS)  & 200 & 15 & 160 & 100 \\ \hline
    \end{tabular}
    \caption{Inference speed at $960 \times 640$. S3G and SG represent S3Gaussian \cite{s3gaussian} and  Street Gaussians \cite{yan2024streetgaussian} respectively.}
        \label{tab:speed}
\end{table}

\section{Editing Examples}
\begin{figure*}
    \centering
    \includegraphics[width=0.9\linewidth]{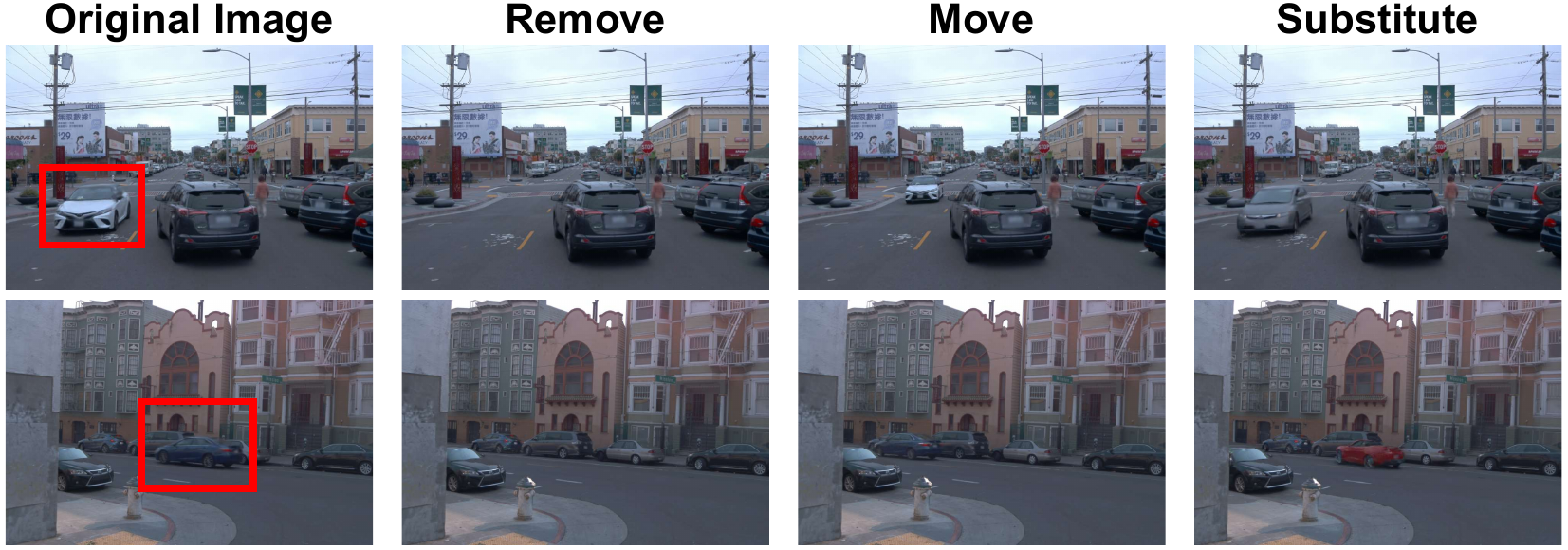}
    \caption{Editing demonstrations on Waymo-NOTR.}
    \label{fig:edit}
\end{figure*}

We provide editing demonstrations in Fig. \ref{fig:edit}.
Gaussians corresponding to cars are associated at initialization and consistently maintained throughout the optimization process.
This enables object editing by directly applying rigid transformations to the corresponding Gaussians.

\section{Choice of 3D tracker}

\begin{table*}[t]
    \centering
    \begin{tabular}{c|c|ccccc}
     & 3D tracker &  PSNR↑ &  SSIM↑ &  LPIPS↓ & DPSNR↑ &  DSSIM↑ \\ \hline
     Street Gaussians & CasTrack \cite{wu2022casa,wu2021castrack} & 25.61 & 0.816 & 0.163 & 21.07 & 0.597 \\
     Street Gaussians & VoxelNext \cite{chen2023voxelnext} & 26.98 & 0.838 & 0.149 & 24.62 & 0.742
    \end{tabular}
    \caption{Ablation study on the choice of 3D tracker for Novel View Synthesis on the Waymo-NOTR dataset.}
    \label{tab:tracker}
\end{table*}

To further illustrate the generalization challenges of 3D trackers, we employ CasTrack \cite{wu2022casa,wu2021castrack} as the 3D tracker for Street Gaussians \cite{yan2024streetgaussian}, using the same detection and tracking algorithm as in the original paper.
Since pretrained nuScenes weights are unavailable, we instead use weights pretrained on KITTI. 
As shown in Table \ref{tab:tracker}, this change leads to a significant performance drop, indicating that merely changing the detection or tracking algorithm does not resolve generalization issues.

\section{Tracking Errors Analysis}
\begin{figure*}
    \centering
    \includegraphics[width=0.99\linewidth]{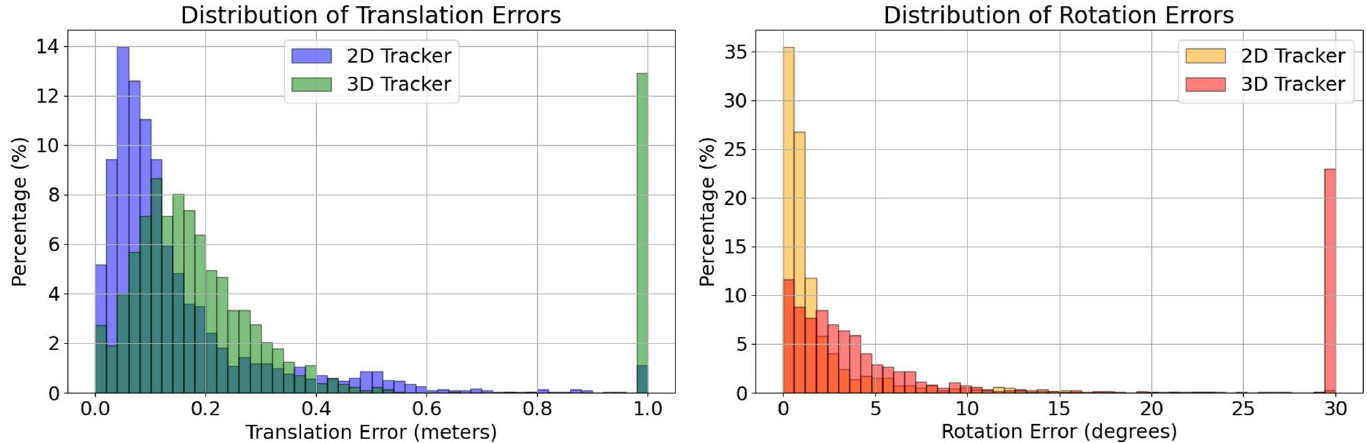}
    \caption{Distributions of tracking errors.}
    \label{fig:track}
\end{figure*}

We evaluate 3D trajectories from 2D and 3D trackers on Waymo-NOTR, measuring translation (Euclidean) and rotation errors (clipped at $1m$ or $30^\circ$ ; with missing detections treated as max error).
Our 2D tracker-based method is more robust. Error distributions are shown in Fig. \ref{fig:track}; a qualitative comparison is shown in Fig. 1 in the main text.

\section{Comparison with Rigid-transformation-based Motion Modeling}
On Waymo-NOTR, we use 3D trajectories computed from 2D tracking to model vehicle motion via rigid transformations, following Street Gaussians (SG) \cite{yan2024streetgaussian} (see Tab. \ref{tab:rigid}). 
While improved trajectories help, SG still underperforms—its joint optimization of pose and Gaussians struggles to converge in frames with large pose noise or missing detections, which in turn degrades the Gaussians across all frames. 
In contrast, we use HexPlane to enforce temporal smoothness by interpolating motion features from a spatio-temporal voxel grid. 
Since we remove trajectory guidance after the first 40\% of training iterations, HexPlane can recover from noisy or missing detections by learning consistent motion from correctly tracked frames.

\begin{table*}[!h]
    \centering
    \begin{tabular}{c|cc|cc}
    Method &  PSNR$\uparrow$ &  SSIM$\uparrow$ &  DPSNR$\uparrow$ &  DSSIM$\uparrow$  \\ \hline
        2D tracker + rigid transformation & 27.69 & 0.850 & 25.04 &  0.758 \\ 
        3D tracker + rigid transformation & 26.98 & 0.838 & 24.62 & 0.742 \\ \hline
        Ours (2D tracker + HexPlane) & 28.85 & 0.867 & 25.58 & 0.779 \\
    \end{tabular}
    \caption{Comparison of rigid-transformation-based motion modeling \cite{yan2024streetgaussian} with our method on the Waymo-NOTR dataset.}
    \label{tab:rigid}
\end{table*}

\section{Details of Loss Functions}
As described in the main paper, the total loss function is expressed as:  
\begin{equation}
\begin{split}
    \mathcal{L} = &\lambda_{\text{rgb}} \mathcal{L}_{\text{rgb}} 
    + \lambda_{\text{ssim}} \mathcal{L}_{\text{ssim}} 
    + \lambda_{\text{depth}} \mathcal{L}_{\text{depth}} 
    + \lambda_{\text{tv}} \mathcal{L}_{\text{tv}} \\
    &+ \lambda_{\text{color-reg}} \mathcal{L}_{\text{color-reg}} 
    + \lambda_{\text{motion}} \mathcal{L}_{\text{motion}}.
\end{split}
\end{equation}

The components of the loss function are detailed below:  
\begin{enumerate}
    \item \textbf{Photometric L1 Loss (\( \mathcal{L}_{\text{rgb}} \)):}  
    This L1 loss measures the photometric difference between the rendered image and the ground truth:  
    \begin{equation}
        \mathcal{L}_{\text{rgb}} = || \mathbf{I}_\text{render} - \mathbf{I}_\text{gt} ||_1,
    \end{equation}
    where $\mathbf{I}_\text{render}$ and $\mathbf{I}_\text{gt}$ represent the rendered and ground truth images, respectively.

    \item \textbf{Structural Similarity Loss (\( \mathcal{L}_{\text{ssim}} \)):}  
    This loss evaluates the structural similarity between $\mathbf{I}_\text{render}$ and $\mathbf{I}_\text{gt}$:  
    \begin{equation}
        \mathcal{L}_{\text{ssim}} = 1.0 - \text{SSIM}(\mathbf{I}_\text{render}, \mathbf{I}_\text{gt}).
    \end{equation}

    \item \textbf{Depth Loss (\( \mathcal{L}_{\text{depth}} \)):}  
    This L1 loss computes the difference between the rendered depth map $\mathbf{D}_\text{render}$ and the ground truth depth map derived from LiDAR data $\mathbf{D}_\text{gt}$:  
    \begin{equation}
        \mathcal{L}_{\text{depth}} = \frac{1}{d} || \mathbf{D}_\text{render} - \mathbf{D}_\text{gt} ||_1,
    \end{equation}
    where $d = 80$ is the predefined maximum depth used for normalization. Depth loss is calculated only for pixels with ground truth depth values between 0.01 and 80 meters.

    \item \textbf{Total Variation Loss (\( \mathcal{L}_{\text{tv}} \)):}  
    A grid-based total variation loss is employed to encourage smooth gradients for HexPlane feature grids, following K-Planes \cite{kplanes_2023}:  
    \begin{equation}
        \mathcal{L}_{\text{tv}} = \operatorname{avg}_{c,i,j} \left( || P_c^{i,j} - P_c^{i-1,j} ||_2^2 + || P_c^{i,j} - P_c^{i,j-1} ||_2^2 \right),
    \end{equation}
    where \(\operatorname{avg}\) denotes the average operator, \(c\) is the plane index, and \(i, j\) are indices on the plane resolution.

    \item \textbf{Color Regularization Loss (\( \mathcal{L}_{\text{color-reg}} \)):}  
    This L1 regularization loss minimizes the predicted color change \(\Delta \mathcal{C}\) for each point to regularize the deformation network:  
    \begin{equation}
        \mathcal{L}_{\text{color-reg}} = \Sigma || \Delta \mathcal{C} ||_1.
    \end{equation}

    \item \textbf{Motion Loss $\mathcal{L}_\text{motion}$:}
    The motion loss is introduced in the main paper as,
    \begin{equation}
    \mathcal{L}_\text{motion} = \operatorname{avg}_{\mathcal{X} \in \mathcal{O}}|\Delta \mathcal{X}_t - (\mathbf{T}_t \mathcal{X} - \mathcal{X}) |,
    \label{eq:motion1}
\end{equation}
where $\mathcal{X}$ is the center position of a Gaussian in object $\mathcal{O}$.
\end{enumerate}

The weights assigned to each loss component are:  
\(\lambda_{\text{rgb}} = 1.0\), \(\lambda_{\text{ssim}} = 0.1\), \(\lambda_{\text{depth}} = 1.0\), \(\lambda_{\text{tv}} = 0.1\), \(\lambda_{\text{color-reg}} = 0.01\), and \(\lambda_{\text{motion}} = 1.0\).

\section{Limitation and Failure Cases}
\begin{figure}
    \centering
    \includegraphics[width=0.99\linewidth]{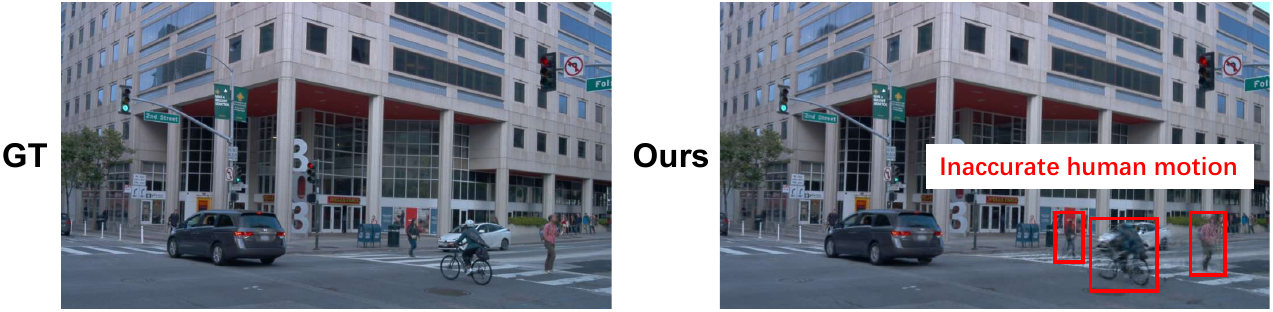}
    \caption{Failure cases}
    \label{fig:failure}
\end{figure}

\begin{itemize}
    \item Our approach primarily focuses on modeling moving vehicles while using 4DGS \cite{4dgs} to model humans without explicit motion guidance. This leads to inaccurate human motion in some cases, which can be improved by incorporating human pose estimation as prior information (see Fig. \ref{fig:failure}). Incorporating human pose estimation techniques \cite{xu2022vitpose, xu2022vitpose+} in future work could enhance human motion modeling.

    \item Dynamic objects beyond humans and vehicles, such as animals and traffic lights, are currently treated as static objects. Further refinement is needed to more effectively distinguish between static and dynamic elements within a scene.

    \item The proposed method requires per-scene optimization. A promising direction for future work is the development of a feed-forward approach for predicting generalizable 3D Gaussians.
\end{itemize}

\end{document}